\documentclass[3p,times]{elsarticle}

\usepackage{ecrc}

\volume{00}

\firstpage{1}

\journalname{Pattern Recognition}

\runauth{}

\jid{procs}

\jnltitlelogo{Pattern Recognition}

\CopyrightLine{2011}{Published by Elsevier Ltd.}

\usepackage{amssymb}
\usepackage[figuresright]{rotating}

\usepackage[pagebackref,breaklinks,colorlinks]{hyperref}

\usepackage{graphicx}
\usepackage{amsmath}
\usepackage{amssymb}
\usepackage{booktabs}

\usepackage{multirow}
\usepackage{xcolor}
\usepackage{amssymb}
\usepackage{bm}
\usepackage{threeparttable}
\usepackage[symbol]{footmisc}

\begin{document}

\begin{frontmatter}



\dochead{Pattern Recognition}
%% e.g. \dochead{17th International Conference on Dynamical Processes in Excited States of Solids}

\title{NCL++: Nested Collaborative Learning for Long-Tailed Visual Recognition}


% \author{Zichang Tan, Jun Li, Jinhao Du, Jun Wan, Zhen Lei, Guodong Guo}

% \address{}
% \address{
% %Manuscript created May, 2022. \\
% %(\textit{Corresponding Author: Guodong Guo}) \\
% Z. Tan, J. Du and G. Guo are with the Institute of Deep Learning, Baidu Research, Beijing, 100085, China.
% %and also with the National Engineering Laboratory for Deep Learning Technology and Application, Beijing 100101, China. 
% E-mail: \{tanzichang, dujinhao, guoguodong01\}@baidu.com.}
% \address{J. Li, J. Wan and Z. Lei are with the National Laboratory of Pattern Recognition (NLPR), Center for Biometrics and Security Research (CBSR), Institute of Automation, Chinese Academy of Sciences (CASIA), Beijing 100190, China, and also with the School of Artificial Intelligence, University of Chinese Academy of Sciences (UCAS), Beijing 100049, China.
% Z. Lei is also with the Centre for Artificial Intelligence and Robotics, Hong Kong Institute of Science \& Innovation, Chinese Academy of Sciences, Hong Kong. E-mail: \{lijun2021, jun.wan\}@ia.ac.cn, zlei@nlpr.ia.ac.cn.}


\renewcommand{\thefootnote}{\fnsymbol{footnote}}
\author[myaddress1]{Zichang Tan\footnotemark[1]}
\author[myaddress2,myaddress3]{Jun Li\footnotemark[1]}
\author[myaddress1]{Jinhao Du}
\author[myaddress2,myaddress3]{Jun Wan}
\author[myaddress2,myaddress3]{Zhen Lei}
\author[myaddress1]{Guodong Guo}
\footnotetext[1]{The first two authors contributed equally to this work}
\address[myaddress1]{Institute of Deep Learning, Baidu Research, Beijing, China}
\address[myaddress2]{Institute of Automation, Chinese Academy of Sciences (CASIA), Beijing, China}
\address[myaddress3]{School of Artificial Intelligence, University of Chinese Academy of Sciences (UCAS), Beijing, China}
%\address[myfourthaddress]{Institute of Automation, Chinese Academy of Sciences, Beijing, China}

\begin{abstract}
Long-tailed visual recognition has received increasing attention in recent years.
Due to the extremely imbalanced data distribution in long-tailed learning,
the learning process shows great uncertainties.
For example, the predictions of different experts on the same image vary
remarkably despite the same training settings.
To alleviate the uncertainty, we propose a Nested Collaborative Learning (NCL++)
which tackles the long-tailed learning problem by a collaborative learning.
To be specific, the collaborative learning consists of two folds,
namely inter-expert collaborative learning (InterCL) and intra-expert collaborative learning (IntraCL).
InterCL learns multiple experts collaboratively and concurrently,
aiming to transfer the knowledge among different experts.
IntraCL is similar to InterCL, but it aims to conduct the collaborative learning on multiple augmented copies of the same image within the single expert.
%One is the inter-expert collaborative learning, where multiple experts are employed and learned concurrently and collaboratively. The other is intra-expert collaborative learning, where the multiple augmented copies of the same image would be input into the same expert.
To achieve the collaborative learning in long-tailed learning, the balanced online distillation is proposed to force the consistent predictions among different experts and augmented copies, which reduces the learning uncertainties. 
Moreover, in order to improve the meticulous distinguishing ability on the confusing categories, we further propose a Hard Category Mining (HCM),
which selects the negative categories with high predicted scores as the hard categories. Then, the collaborative learning is formulated in a nested way, in which the learning is conducted on not just all categories from a full perspective but some hard categories from a partial perspective. Extensive experiments manifest the superiority of our method with outperforming the state-of-the-art whether with using a single model or an ensemble. 
%Code will be made publicly.
The code is available at \url{https://github.com/Bazinga699/NCL}.
\end{abstract}

\begin{keyword}
Long-tailed visual recognition, collaborative learning, online distillation, deep learning.
\end{keyword}

\end{frontmatter}

% \linenumbers

\section{Introduction}
\label{sec:intro}

\begin{figure}[h]
\centering
    {\includegraphics[width=0.7\linewidth]{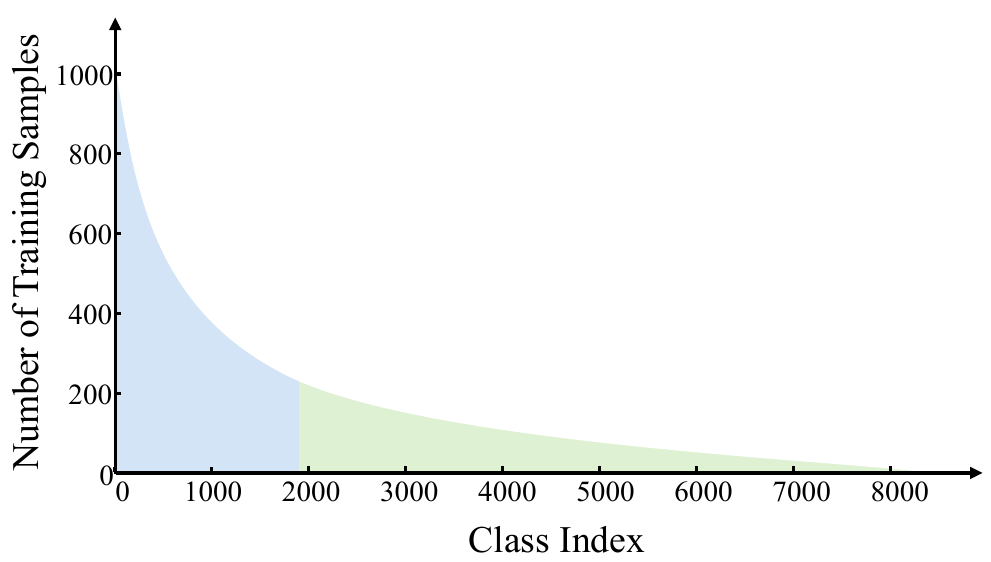}}
    \caption{
    \iffalse
    An illustration of long-tailed data distribution.
    %on iNaturalist 2018 dataset~\cite{van2018inaturalist}. 
    In the distribution, few head classes occupy most of the data while 
    many tail classes have only a few samples.
    The proposed model aims to alleviate the learning uncertainties, which are shown in the following Fig.~\ref{fig_uncertainty}. 
    %The model shows great uncertainty that are shown in the following Fig.~\ref{fig_uncertainty}. 
    The uncertainty mainly comes from two folds.
    One is that, for two experts with the same configuration and the same training settings, they still vary greatly on predictions with respect to the same input image.
    The other is that  
    the same expert performs differently
    for the same image with different augmentations.
    Such phenomenon is more pronounced in the tail category of long-tailed distributions (see details in Fig.~\ref{fig_uncertainty}). 
    %The learning uncertainty exists not only among different experts, but also in the same expert which takes different augmented copies of the same image as the input. Such phenomenon is more pronounced in the tail category of long-tailed distributions. We expand the exploration of uncertainty in long-tailed learning, encouraging the network to learn more essential and invariant features under diverse conditions.   
    \fi
    An illustration of long-tailed data distribution.
    In the distribution, few head classes occupy most of the data while many tail classes have only a few samples. 
    %The proposed method aims to alleviate the learning uncertainties from two aspects. One is that the same expert performs differently for the same image with different augmentations. The other is that, for two experts with the same configuration and the same training settings, they still vary greatly on predictions with respect to the same input image. To alleviate such uncertainties, we employ the collaborative learning to encourage the network to learn essential and invariant features under diverse conditions. 
}
\label{fig_intro}
\end{figure}

In recent years, deep neural networks have achieved resounding success in various visual tasks, e.g., image classification~\cite{he2016deep,hu2018squeeze},  object detection~\cite{ren2015faster,zhang2018single}, semantic segmentation~\cite{zhao2017pyramid,fu2019dual} and so on. Despite the advances in deep technologies and computing capability,
the huge success also highly depends on large well-designed datasets of having a roughly balanced distribution,
such as ImageNet~\cite{deng2009imagenet}, MS COCO~\cite{lin2014microsoft} and Places~\cite{zhou2017places}. 
%However, those datsets are well-designed with human efforts 
%and have a roughly balanced distribution
%with each category containing adequate samples.
This differs notably
%is very different 
from real-world datasets, which usually exhibits 
long-tailed data distributions~\cite{wang2017learning,liu2019large}
where few head classes occupy most of the data while many tail classes only have few samples as shown in Fig.~\ref{fig_intro}.
In such scenarios, the model is easily dominated by those few head classes,
whereas low accuracy rates are usually achieved for many other tail classes.
Undoubtedly, the long-tailed characteristics challenge deep visual recognition,
and also immensely hinder the practical use of deep models.

In long-tailed visual recognition, most early works focus on designing the class re-balancing strategies~\cite{he2009learning,buda2018systematic,cui2019class,huang2016learning,ren2018learning,wang2017learning} and decoupled learning~\cite{kang2019decoupling,cao2019learning}.
Those methods can accomplish some accuracy improvements but 
still cannot deal with the long-tailed class imbalance problem well.
For example, class re-balancing methods are often confronted with risk of overfitting. More recent efforts aim to improve the long-tailed learning by using multiple experts~\cite{xiang2020learning,wang2020long,li2020overcoming,cai2021ace,zhang2021test}.
The multi-expert algorithms follow a straightforward idea of making them diverse from each other.
To achieve this, some works~\cite{xiang2020learning,zhang2021test} force different experts to focus on different aspects. For example, LFME~\cite{xiang2020learning} formulates a network with three experts
and it forces each expert to learn samples from one of head, middle and tail classes.
Besides, there are some works~\cite{wang2020long} adopt some constraint losses to diversify the multiple experts.
For example, RIDE~\cite{wang2020long} proposes a distribution-aware diversity loss to achieve this.
%which means that different experts focus on different aspects  and each of them benefits from the specialization in the dominating part.
%For example, LFME~\cite{xiang2020learning} formulates a network with three experts and it forces each expert learn samples from one of head, middle and tail classes.
%% Multi-expert framework usually can achieve satisfactory recognition accuracy on long-tailed recognition since each of those experts is well learned on its own field.
%% We also deal with the long-tailed class distribution following 
%% the multi-expert scheme due to its advantages.
%Previous multi-expert methods~\cite{wang2020long,xiang2020learning,cai2021ace}, however, force each expert to only learn the knowledge in a specific area, and there is a lack of cooperation among them.
The previous multi-expert methods can achieve good performance mainly due to the diversity among them. Also because of this, these methods obtain the predictions through the ensemble rather than a single expert due to that only employing a single expert hardly achieves reliable recognition.
%hardly achieve promising predictions through a single expert, but an ensemble of them.

\begin{figure}[h]
\centering
    {\includegraphics[width=0.6\linewidth]{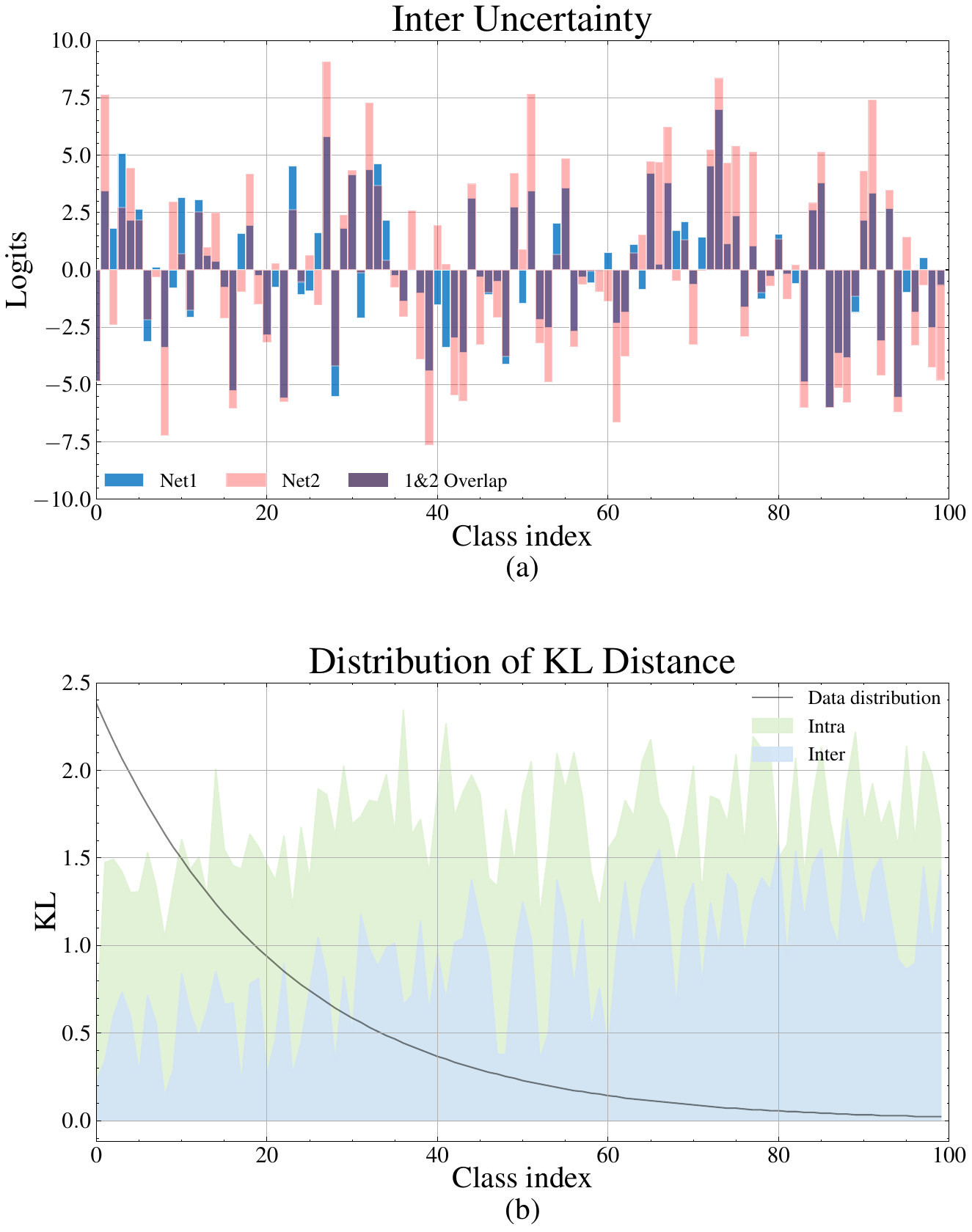}}
    \caption{
    (a) The Kullback–Leibler (KL) distance calculated from two aspects.
    The green color indicates the KL distance of two experts with taking the same image as the input. The blue color indicates the KL distance of an expert with taking two augmented image copies with respect to the same image as the inputs.
    (b) An illustration of the predictions produced by two experts with respect to the same input. The two experts have the same structure and are trained with the same settings.
    The analysis is conducted on CIFAR100-LT dataset with an Imbalanced Factor (IF) of 100.
    The predictions are visualized on the basis of a random selected example,
    and the KL distance is computed based on the whole test set and then 
    the average results of each category are counted and reported.
    The predictions differ largely from each other between different networks and different augmented images. Bested viewed in color.
    %The comparisons of model outputs (logits) and Kullback–Leibler (KL) distance from two aspects, i.e., the same image but predicted by different networks and the same network with taking different augmentations of the same image as the input. Analysis is conducted on CIFAR100-LT dataset with Imbalanced Factor (IF) of 100. The logits are visualized on the basis of a random selected example, and the KL distance is computed based on the whole test set and then the average results of each category are counted and reported. The predictions differ largely from each other between different networks and different augmented images. 
    % Although the employed two networks have the same network structure and training settings,
    % their predictions differ largely from each other especially in tail classes.Bested viewed in color.
}
\label{fig_uncertainty}
\end{figure}

\iffalse
However, we found that different models have inherent diversity even though 
they have the same network structure and training settings.
As shown in Fig.~\ref{fig_uncertainty}, we visualize the predictive distributions of two different experts with the same structure and training settings on a randomly selected sample. The two experts vary greatly on predictions especially in tail classes, which signifies the great uncertainty in the learning process as well as the diversity in models.
In addition, the predictions of the same expert with taking different augmentations of the same image are also visualized in~\ref{fig_uncertainty}.
Similar uncertainties still remain.
According to previous works~\cite{xiang2020learning,wang2020long,li2020overcoming,cai2021ace,zhang2021test}, 
a simple ensemble like averaging the predictions could be employed to reduce the uncertainty in the prediction stage,
which can achieve a certain performance improvement. 
However, an ensemble of multiple experts also brings huge amount of calculation and parameters,
which limits its practical use. This brings up a question, \textit{can we utilize these uncertainties to improve the performance of a single model?}
\fi

However, we found that there are great uncertainties in the long-tailed learning. The learning uncertainties mainly come from two aspects. One is that the same expert performs differently for the same image with different augmentations. The other is that, for two experts with the same configuration and the same training settings, they still vary greatly on predictions with respect to the same input image. To analyze this, we visualize the Kullback-Leibler (KL) distance\footnote{https://en.wikipedia.org/wiki/Kullback-Leibler\_divergence}
over the predictions generated by two experts with the same training settings, and the predictions generated by two augmented image copies with respect to the same expert. As shown in Fig.~\ref{fig_uncertainty} (b), the green color shows the KL distance distribution for two different experts of the same structure and training settings with taking the same image as the input, and the blue color shows that of the same expert with taking different augmentations of the same image. Specifically, to analyze the uncertainties across all categories, the Kullback-Leibler (KL) distance is specifically calculated for all categories with reporting the average distance over all images for each class.
As we can see, the predictions vary greatly especially in tail classes, which signifies the great uncertainty in the learning process as well as the diversity in models.
%we visualize the predictive distributions of two different experts with the same structure and training settings on a randomly selected sample. 
%The two experts vary greatly on predictions especially in tail classes, which signifies the great uncertainty in the learning process as well as the diversity in models.
%In addition, the predictions of the same expert with taking different augmentations of the same image are also visualized in~\ref{fig_uncertainty}.
%Similar uncertainties still remain.
To show the prediction differences in details,
we visualize the detailed predictions of the two experts with the same structure and training settings on a randomly selected sample (corresponding to the green color in Fig.~\ref{fig_uncertainty} (b)).
According to previous works~\cite{xiang2020learning,wang2020long,li2020overcoming,cai2021ace,zhang2021test}, 
a simple ensemble like averaging the predictions could be employed to reduce the uncertainty in the prediction stage,
which can achieve a certain performance improvement. 
However, an ensemble of multiple experts also brings huge amount of calculation and parameters, which limits its practical use. This brings up a question, \textit{can we utilize these uncertainties to improve the performance of single model?}

The goal of this paper is to enhance the performance of a single model via 
reducing the learning uncertainty
by the collaborative learning.
%To be specific, each expert in the multi-expert can learn the extra know.
In the multi-expert framework, owing to the uncertainty in the learning process (e.g., the parameter initialization), different experts capture various knowledge.
Besides, the learning uncertainty of the model itself can also lead to various predictions
%that the predictions will change greatly 
when adding some noises or augmentations to the input.
To reduce the uncertainty in the learning process, the Inter-expert Collaborative Learning (InterCL) is employed to collaboratively learn multiple experts concurrently, where each expert serves as a teacher to teach others and also as a student to learn extra knowledge from others. 
%The collaborative learning on multiple experts allows each expert to be teacher to teach others and also to be student to learn extra knowledge from others. 
%To achieve this, the inter-expert distillation is employed to learn all experts .
In this way, all experts tend to learn the consistent and right knowledge. 
In addition to the uncertainty among different experts, the predictions of the same expert with taking different augmentations of the same image as the input also show great uncertainties. Grounded in this, we augment the image by multiple times and then input them to the same expert. Later, the Intra-expert Collaborative Learning (IntraCL) is further employed to distill knowledge among different augmentations
and also reduce the learning uncertainty among them.
Specifically, we employ the online distillation to achieve the goal of collaborative learning. In order to better cope with the long-tailed data distribution,
we further formulate the online distillation in a balanced way, i.e., balanced online distillation. Specifically, in the proposed balanced online distillation, 
the contributions of tail classes would be strengthened while that of head classes would be suppressed.

Previous works~\cite{guo2020online,lan2018knowledge,zhang2018deep} simply conduct the classification and distillation from a full perspective on all categories. It helps the network to obtain the global discrimination on all categories,
but lacks of meticulous distinguishing ability on the confusing categories.
%which is inherently important for visual recognition.
%However, the meticulous distinguishing ability is inherently important for visual recognition.
In the classification problem, 
%compared with the easy negative categories with low predictive scores, 
the network should pay attention to the hard negative categories with high predictive scores 
rather than simply treating all categories equally,
%than the easy negative categories with low predictive scores, 
which helps to reduce the confusion between the target category and the confusing categories.
To achieve this, we first propose a Hard Category Mining (HCM) to select the hard categories for each sample, in which the hard category is defined as the negative category with a high predictive score.
Then, we formulate a nested learning for the supervised individual learning of a single network as well as the collaborative learning (including IntraCL and InterCL).
To the specific, the nested learning conducts the supervised learning or distillation from two perspectives, where one is the full perspective on all categories and the other is the partial perspective only on the selected hard categories. 
In this way, not only the global discriminative capability  but also the meticulous distinguishing ability can be captured. 
%Furthermore, we propose a class-wise regularization to force the network output consistent distributions of the samples of the same class, which can further reduce the intra-class variations.

The proposed method utilizes the collaborative learning for long-tailed visual recognition. The proposed collaborative learning is two folds,
where one is inter-expert collaborative learning and the other is intra-expert collaborative learning. The collaborative learning allows the knowledge transferring 
among different experts and image augmentations, which aims to 
reduce the learning uncertainties and promotes each expert model to achieve better performance. Our method is inherently a multi-expert framework. 
However, unlike previous works that need an ensemble of all experts to obtain the final predictions, our method can achieve the state-of-the-art performance only based on a single expert (better performance can be achieved based on the ensemble of all). 
This is because we collaboratively learn multiple experts by reducing the diversity of models rather than increasing their diversity as in previous works~\cite{xiang2020learning,wang2020long,li2020overcoming,cai2021ace,zhang2021test}. 
%our approach can achieve this by using an ensemble as well as only a single network. Experiments will show that even with an expert, it can achieve very good performance, even comparable to the performance of the ensemble.
Our contributions can be summarized as follows:
%\vspace{-0.15cm}
\begin{itemize}
\setlength{\itemsep}{1.0pt}
  \item We propose a Nested Collaborative Learning (NCL++) to improve long-tailed visual recognition via 
  collaborative learning. The collaborative learning consists of two folds, 
  namely intra-expert and inter-expert collaborative learning,
  which allow the knowledge transferring among different image augmented copies and experts, respectively.
  To our best knowledge, 
  \textbf{
  it is the first work to adopt the collaborative learning to 
  address the problem of long-tailed visual recognition}.
  %where one is the intra-expert collaborative learning that learns .
  %to collaboratively learn multiple experts concurrently, which allows each expert model to learn extra knowledge from others.
  %Hard Category Focus (HCF), which facilitates the network to focus on categories that are hard to be distinguished.
  \item We propose a Nested Feature Learning (NFL) to conduct the learning from both a full perspective on all categories and a partial perspective of focusing on hard categories. This helps the model to capture meticulous distinguishing ability on confusing categories.
  \item We propose a Hard Category Mining (HCM) to select hard categories for each sample.
  %distinguish the sample
  %from the hard negative categories.
  %select hard categories for ,
  %and helps greatly reduce the confusion with hard categories.
  %\item We propose to enhance feature learning via self-supervision, and further employ it to assist multi-expert collaborative learning.
  \item The proposed method gains significant performance over the state-of-the-art on five popular datasets including CIFAR-10/100-LT,  Places-LT, ImageNet-LT and iNaturalist 2018.
\end{itemize}

\section{Related Work}
\label{sec:related}

%In the following, we review related works in long-tailed visual recognition and knowledge distillation.

\subsection{Long-tailed visual recognition.}
To alleviate the long-tailed class imbalance,
lots of studies ~\cite{zhou2020bbn,cao2020domain,xiang2020learning,ren2020balanced,zhang2021bag,wang2020long} 
have been conducted in recent years. The existing methods for long-tailed visual recognition 
can be roughly divided into three categories: class-rebalancing~\cite{he2009learning,buda2018systematic,cui2019class,huang2016learning,ren2018learning,wang2017learning}, 
multi-stage training~\cite{kang2019decoupling,cao2019learning} and multi-expert methods~\cite{xiang2020learning,wang2020long,li2020overcoming,cai2021ace,zhang2021test}.
In the following, we will review the methods in the above three categories.

Class re-balancing, which aims to re-balance the contribution of each class during training,
is a classic and  widely used method for long-tailed learning.
Usually, there are two types for class re-balancing.
The first one is data re-sampling~\cite{chawla2002smote,kang2019decoupling,wang2020devil,zhang2021learning,zhang2021bag}, including over-sampling~\cite{chawla2002smote}, 
under-sampling, square-root sampling~\cite{kang2019decoupling} and progressively-balanced sampling~\cite{kang2019decoupling}. 
The goal of data re-sampling is to increase the sampling probability of samples in tail classes
while weakening that of samples in head classes.
%More specifically, class re-balancing consists of data re-sampling~\cite{chawla2002smote,kang2019decoupling,wang2020devil,zhang2021learning,zhang2021bag}, (e.g., over-sampling～\cite{chawla2002smote}, under-sampling, square-root sampling~\cite{kang2019decoupling} and progressively-balanced sampling~\cite{kang2019decoupling}), 
The second one is loss re-weighting~\cite{lin2017focal,wang2021seesaw,ren2020balanced,tan2020equalization,zhao2018adaptive,zhang2021bag},
that is, re-weighting of loss function by the numbers of different classes.
The popular methods in loss re-weighting include Focal loss~\cite{lin2017focal}, Seesaw loss~\cite{wang2021seesaw}, Balanced Softmax Cross-Entropy (BSCE)~\cite{ren2020balanced}, Class-Dependent Temperatures (CDT)~\cite{ye2020identifying}, LDAM loss~\cite{cao2019learning} and Equalization loss~\cite{tan2020equalization}.
%and post-adjusting the model logits via label frequencies~\cite{menon2020long,zhang2021distribution}.
%(e.g.,  logit adjustment~\cite{menon2020long} and DisAlign~\cite{zhang2021distribution}).
Class re-balancing improves the overall performance but usually at the sacrifice of the accuracy on head classes.

Multi-stage training methods divide the training process into several stages~\cite{hong2021disentangling,menon2020long,zhang2021distribution,kang2019decoupling,li2021self}.
For example, Kang et al.~\cite{kang2019decoupling} decouple the training procedure
into representation learning and classifier learning,
where the representation learning adopts the instance-balanced sampling to learn a good feature extractor and the classifier learning adopts the class-balanced sampling to re-adjust the classifier. Besides, some other works~\cite{menon2020long,zhang2021distribution,hong2021disentangling,zhao2021adaptive} tend to improve performance via a post-process of shifting model logits.
Li et al.~\cite{li2021self} propose a four-stage training strategy on basis of 
knowledge distillation, self-supervision and decoupled learning.
To be specific, the network is first trained with classification and self-supervised losses,
and then the feature extractor is frozen and the classifier is re-adjusted by the class-balanced sampling. Later, an additional network is employed to distill the knowledge from the previously trained network, and finally, the network is further trained by re-adjusting the classifier.
As we can see, multi-stage training methods may rely on heuristic design.

More recently, multi-expert frameworks~\cite{xiang2020learning,zhou2020bbn,wang2020long,zhang2021test,cai2021ace,li2020overcoming,wang2020devil,li2021trustworthy} receive increasing concern, e.g.,  Learning From Multiple Expert (LFME)~\cite{xiang2020learning}, Bilateral-Branch Network (BBN)~\cite{zhou2020bbn}, RoutIng Diverse Experts (RIDE)~\cite{wang2020long}, Test-time Aggregating Diverse Experts (TADE)~\cite{zhang2021test} and Ally Complementary Experts (ACE)~\cite{cai2021ace}.
For example, LFME~\cite{xiang2020learning} formulates three experts with each corresponding to one of head, medium and tail classes.
RIDE~\cite{wang2020long} proposes a distribution-aware diversity loss to the multi-expert network
and it encourages the experts to be diverse from each other.
Multi-expert methods indeed improve the recognition accuracy for long-tailed learning, 
but those methods still need to be further exploited.
For example, most current multi-expert methods employ different models
to learn knowledge from different aspects, while the mutual supervision among them is deficient.
Moreover, they often employ an ensemble to produce predictions, which leads to an increase in complexity.

\subsection{Knowledge distillation.}
Knowledge distillation is a prevalent technology in knowledge transferring.
One typical manner of knowledge distillation is teacher-student learning~\cite{hinton2015distilling,furlanello2018born},
which transfers knowledge from a large teacher model to a small student model.
Current methods in knowledge distillation can be divided into three categories:
offline distillation~\cite{hinton2015distilling,romero2014fitnets,passalis2018learning,li2020few,zagoruyko2016paying}, online distillation~\cite{zhang2018deep,guo2020online,chen2020online,dvornik2019diversity,chen2020online,bhat2021distill} 
and self-distillation~\cite{zhang2020self,zhang2019your,hou2019learning,phuong2019distillation,yuan2019revisit}.
Early methods~\cite{hinton2015distilling,passalis2018learning} often adopt an offline learning strategy, which transfers the knowledge from a pretrained teacher model to a student model.
Most works~\cite{hinton2015distilling,passalis2018learning,mirzadeh2020improved} distill the knowledge from the output distributions, 
while some works achieve the knowledge transferring by
matching feature representations~\cite{romero2014fitnets} or attention maps~\cite{zagoruyko2016paying}.
The offline distillation is very popular in the early stage.
However, the offline way only considers transferring the knowledge from the teacher to the student, and therefore, the teacher normally should be a more complex high-capacity model than the student. 
%Besides, the training process may be cumbersome and time-consuming.
In recent years, knowledge distillation has been extended to an online way~\cite{zhang2018deep,guo2020online,chen2020online,dvornik2019diversity,chen2020online,walawalkar2020online}, where the whole knowledge distillation is conducted in an one-phase and end-to-end training scheme. For example, in Deep Mutual Learning~\cite{zhang2018deep}, any one model can be a student
and can distill knowledge from all other models.
%Guo et al.~\cite{guo2020online} propose to use an ensemble of soft logits to guide the learning.
Zhu et al.~\cite{lan2018knowledge} propose a multi-branch architecture with treating each branch as a student to further reduce computational cost. 
%Online distillation is an efficient way to collaboratively learn multiple models, and facilitates the knowledge transferred among them.
Compared with offline distillation, online distillation is more efficient by taking the learning in an one-phase end-to-end scheme.
For self-distillation~\cite{zhang2020self,zhang2019your,hou2019learning,phuong2019distillation,yuan2019revisit}, it can be regarded as a special case in online distillation, 
where the teacher and the student refer to the same network.
In other words, self-distillation means that the model distills the knowledge from itself.
For example, Zhang et al.~\cite{zhang2019your} divide the network into several sections according to their depth, and allow the low sections to distill the knowledge from high sections.
%Similar idea is also adopted in the work~\cite{hou2019learning}, where the preceding layers mimic the attention maps of the deeper layers. 
Our work aims to reduce the predictive uncertainty in long-tailed learning by
online distillation and self-distillation. 
However, different from previous works, we formulate the distillation in a nested way to focus on all categories as well as some important categories, which facilitates the network to obtain the meticulous distinguishing ability.

\begin{figure*}[t!]
\centering
    {\includegraphics[width=1.0\linewidth]{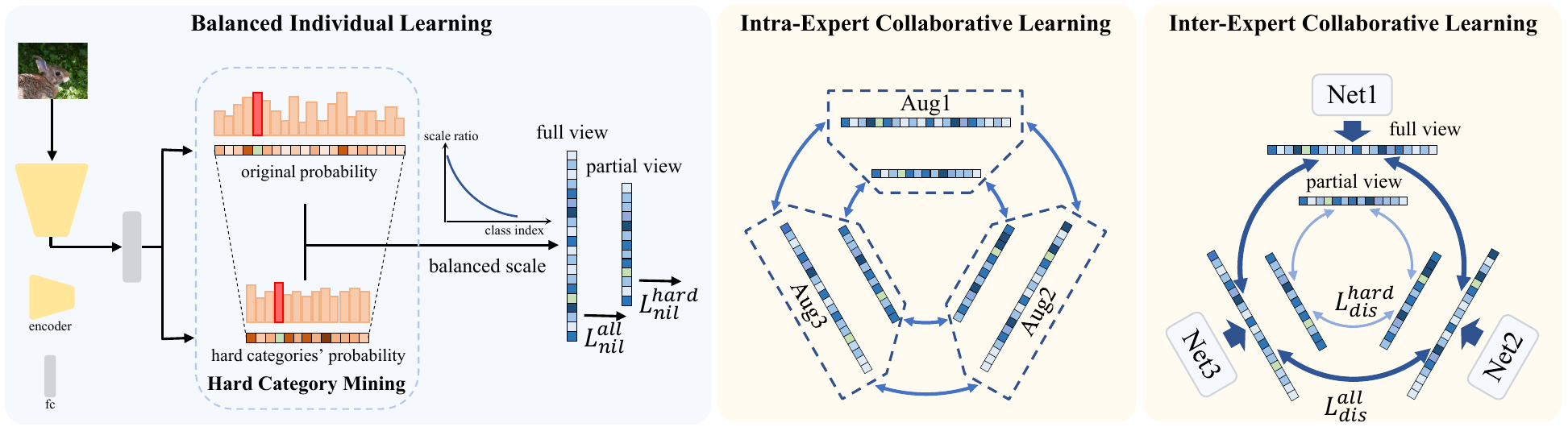}}
    \caption{
    An illustration of our proposed NCL++ of containing three experts and three augmented image copies 
    for conducting collaborative learning. 
    The proposed NCL++ contains five core components, namely Balanced Individual Learning (BIL),
    Intra-expert Collaborative Learning (IntraCL), Inter-expert Collaborative Learning (InterCL),
    Hard Category Mining (HCM) and Nested Feature Learning. The BIL aims to enhance discriminative ability of a single expert. Both IntraCL and InterCL aim to reduce the learning uncertainty by collaborative learning and thus improve the discriminative capability. For the proposed HCM, 
    it selects the hard negative categories, which are used for improving the meticulous distinguishing ability of the model. Based on the selected hard categories, the NFL is further employed to learning features from a global view on all categories and also a partial view on hard categories.
    %and NBOD allows knowledge transferring among multiple experts. NIL conducts the supervised learning from both a full and a partial views, which focus on all categories and some hard categories, respectively. Similarly, NBOD conducts the knowledge distillation also from both a full and a partial views. The contrastive loss is calculated by using a extra momentum encoder and MLP layers, which can be removed in evaluation. Probabilities employed in NIL and NBOD are balanced according to the data distribution.
    }
    \label{fig_framework}
\end{figure*}

\section{Methodology}
\label{sec:method}
The proposed NCL++ aims to adopt the collaborative learning to reduce the learning uncertainties in long-tailed visual recognition as
shown in Fig.~\ref{fig_framework}. We first introduce the Balanced Individual Learning (BIL),
Intra-expert Collaborative Learning (IntraCL) and Inter-expert Collaborative Learning (InterCL).
Then, we present the Hard Category Mining (HCM) of selecting hard categories, 
and further give a detailed introduction of the Nested Feature Learning (NFL).
%collaboratively and concurrently learn multiple experts together as shown in Fig.~\ref{fig_framework}.
%In the following, we first introduce the preliminaries, and then present Hard Category Mining (HCM),  Nested Individual Learning (NIL) and Nested Balanced Online Distillation (NBOD) and self-supervision part. 
Finally, we show the overall loss of how to aggregate them together.

\subsection{Balanced Individual Learning}
We denote the training set with $n$ samples as ${ {\mathcal D}} = \{ {\bf x}_i, y_i \}$,
where ${\bf x}_i$ indicates the $i$-th image sample and $y_i$ denotes the corresponding label.
Assume a total of $K$ experts are employed and the $k$-th expert model
is parameterized with ${\bm \theta}_k$. 
Given image ${\bf x}_i$, the predicted probability of class-$j$ in the $k$-th expert is computed as:
\begin{equation}\label{ce_prob}
     {\tilde {\bf p}}_j({\bf x}_i;{\bm \theta}_k) = \frac{exp({ {z}_{ij}^k }) }{\sum_{l=1}^C exp({{z}_{il}^k }) }
\end{equation}
where $z_{ij}^k$ is the model's class-$j$ output and $C$ is the number of classes.
This is a widely used way to compute 
the predicted probability, and some losses like Cross Entropy (CE) loss 
is computed based on it.
However, this way does not consider the data distribution,
and is not suitable for long-tailed visual recognition,
where a vanilla model based on ${\tilde {\bf p}}({\bf x}_i;{\bm \theta}_k)$
would be largely dominated by head classes.
Therefore, some researchers~\cite{ren2020balanced} proposed to 
compute predicted probability of class-$j$ in a balanced way:
\begin{equation}\label{bsce_prob}
   {\bf p}_{j} ({\bf x}_i;{\bm \theta}_k) = \frac{n_{j} exp({ {z}_{ij}^k }) }{\sum_{l=1}^C n_l exp({{z}_{il}^k  })  }  
\end{equation}
where $n_j$ is the total number of samples of class $j$.
In this way, contributions of tail classes are strengthened while contributions of head classes are suppressed.
Based on such balanced probabilities, Ren et al.~\cite{ren2020balanced}
further proposed a Balanced Softmax Cross Entropy (BSCE) loss to alleviate long-tailed class imbalance
in model training.
We also adopt the BSCE to conduct the individual learning for each expert, 
which ensures that each one can achieve the strong discrimination ability.
Mathematically, the loss of the balanced individual learning for expert-$k$ can be denoted as:
\begin{equation}\label{eq_bil}
   L_{bil}^{\cal G} = -\textstyle{\sum\nolimits_k} log( { {\bf p}}_{y_i} ({\bf x}_i;{\bm {\bm \theta} }_k) )
\end{equation}
where ${\bm \theta}_k$ indicates the parameters of the $k$-th expert.
The superscript ${\cal G}$ indicates the loss is conducted from a global view on all categories.
%However, only BSCE loss is still not enough, where the uncertainty in training still can not be eliminated.

\subsection{Inter-Expert Collaborative Learning}
To collaboratively learn multiple experts from each other,
we employ the Inter-expert Collaborative Learning (InterCL)
to allow each model to learn extra knowledge from others.
%Different form the IntraBD which distills the knowledge among different augmented copies within the same expert, the proposed InterBD aims to transfer the knowledge among different experts.
We employ the knowledge distillation to allow the knowledge transferring among them,
and following previous works~\cite{zhang2018deep,guo2020online},
the Kullback Leibler (KL) divergence is employed to calculate the loss, which can be denoted as:
\begin{equation}\label{inter_dis}
   L_{inter}^{\cal G} = 
   \frac{2}{K(K-1)} {\sum_{k}^{K} \sum_{q \neq k}^{K} }
   KL( {\bf p} ({\bf x}_i;{\bm \theta}_k) || {\bf p} ({\bf x}_i;{\bm \theta}_q) )
\end{equation}
where $K$ denotes the number of experts
and $KL({\bf p}  ||  {\bf q})$ indicates the KL divergence between the distributions $\bf p$ and $\bf q$,
which can be specifically denoted as
$ KL(\mathbf{p} || \mathbf{q}) =   {\textstyle \sum_{j} \mathbf{p}_j  log( \frac{\mathbf{p}_j}{\mathbf{q}_j}  )} $.
Note that we adopt the balanced probabilities ${\bf p}$ for distillation,
which increases the contributions of tail classes while suppressing that of head classes.
%Note that different previous distillation~\cite{zhang2018deep,guo2020online}, a balanced distillation is employed here to increase the distributions .
In this way, each expert can be the teacher to teach others,
and each expert also can be a student to learn knowledge from others.
Such collaborative learning aims to reduce the prediction uncertainties among multiple experts through minimizing the KL divergence.
Note that the distillation is also set in a balanced way,
which helps the model to learn better for long-tailed visual recognition.
%The proposed InterBD is also could be *** with IntraBD as we have mentioned above, ***

\subsection{Intra-Expert Collaborative Learning}
The learning uncertainty exists not only in different experts, 
but also in the same expert which takes different augmented images as the input.
In other words, given an input image, if we augment it by several times and then feed them into the network,
%For an input image, if it is augmented by several times and we feed all augmented versions into the network,
the corresponding outputs also vary greatly.
Inspired by this, we further propose the Intra-Expert Collaborative Learning (IntraCL), 
which helps the network to alleviate the uncertainty within the expert.
To be specific, for an input image ${\bf x}_i$, assume it will be augmented for $T$ times (like using RandAugment~\cite{cubuk2020randaugment} or other data augmentation strategies), 
and the augmented images are denoted as $\{ {\bf x}_i^t \}_{t=1}^{T}$.
All the augmented images are fed to the expert-$k$ and produce the corresponding balanced probabilities $\{ {\bf p}({\bf x}_i^t;{\bm \theta}_k)\}_{t=1}^{T}$. 
Then, the knowledge distillation is performed to guide the network 
learning reliable and consistent outputs for different augmented copies.
Moreover, the distillation also allows the knowledge transferring among the different augmented copies,
which helps them to learn extra knowledge from each other.
%To be specific, we follow previous works~\cite{zhang2018deep,guo2020online} to employ
%the Kullback Leibler (KL) divergence to calculate the loss of knowledge distillation.
Similarly, KL divergence is adopted for loss calculation,
and the corresponding loss of expert-$k$ can be represented as:
%alleviate such uncertainty in the learning process, 
%To be specific, the input image ${\bf x}_i$ are augmented by many times and then fed to the network. and.
\begin{equation}\label{eq_intra_k}
   L_{intra}^{{\cal G}, k} = 
   \frac{2}{T(T-1)} {\sum_{t}^{T} \sum_{\tau \neq t}^{T} }
   KL( {\bf p} ({\bf x}_i^{t};{\bm \theta}_k) || {\bf p} ({\bf x}_i^{\tau};{\bm \theta}_k) )
\end{equation}
Since the proposed method is a multi-expert framework, the proposed IntraCL would be applied to all experts.
The summed IntraCL loss on all experts is represented as:
\begin{equation}\label{eq_intra_sum}
L_{intra}^{{\cal G}} =  { \textstyle \sum_{k} L_{intra}^{{\cal G}, k} } 
\end{equation}

\begin{figure}[t]
\centering
    {\includegraphics[width=0.5\linewidth]{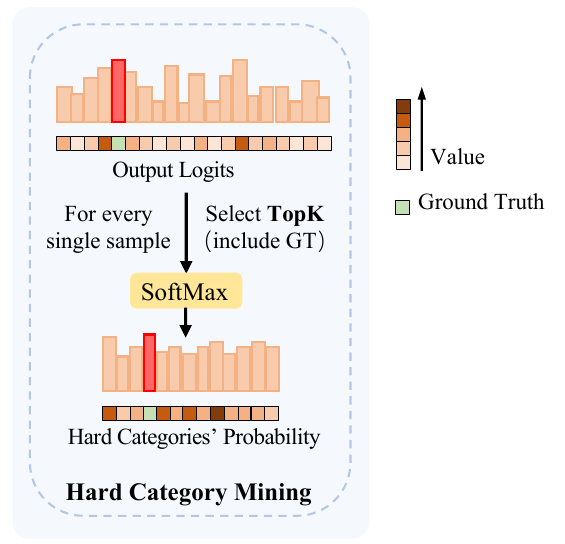}}
    %\vspace{-0.6cm}
    \caption{
    An illustration of the proposed HCM module.
    }
    \label{fig_hcm}
\end{figure}

\subsection{Hard Category Mining}
In representation learning, one well-known and effective strategy to boost performance
is Hard Example Mining (HEM)~\cite{hermans2017defense}.
HEM claims that different examples have different importance, specifically, hard samples are more important than easy samples.
Therefore, HEM selects hard samples for training,
while discarding easy samples which contribute very little and are even detrimental to features learning. 
However, directly applying HEM to long-tailed visual recognition
may distort the data distribution and make it more skewed in long-tailed learning.
Although it can not be directly used for long-tailed learning, 
it still gives us an important inspiration, that is, 
\textit{do different categories have different degrees of importance in the training process?}

When conducting the classification,
the image may be wrongly classified to some confusing categories, 
like the categories similar to the target category.
Therefore, the network should pay close attention to those confusing categories.
The confusing category is also called as the hard category in this paper.
To be specific, the hard categories are defined as the categories that are not the ground-truth category but with high predicted scores. 
With the help of focusing on hard categories, the ability of distinguishing the sample from the negative hard categories can be explicitly improved.
As shown in Fig.~\ref{fig_hcm}, we propose a Hard Category Mining (HCM) to select the hard categories and the target category out, 
and re-calculate the probabilities over those categories, which are used to be trained in the NFL to increase the ability of distinguishing the sample from the negative hard categories.
To be specific, assume we have $C$ categories in total and suppose $C_{hard}$ categories are selected to focus on. For the sample ${\bf x}_i$ and expert $k$, the corresponding set ${\bm \Psi}_i^k$ containing the outputs of selected categories is denoted as:
%the model outputs ${\bm \Psi}_i$ of selected categories is denoted as:
\begin{equation}\label{hard_select}
   {\bm \Psi}_i^k = TopHard\{ z_{ij}^k | j \neq y_i \} \cup \{z_{iy_i}^k \}
\end{equation}
where $TopHard$ means selecting $C_{hard}$ examples with largest values.
In order to better adapt to long-tailed learning,
we compute the  probabilities of the selected categories in a balanced way,
which is shown as: 
\begin{equation}\label{hard_p}
   {\bf p}^* ({\bf x}_i;{\bm \theta}_k) = \{  
   \frac{n_j exp( { {z}_{ij}^k }) }{\sum_{z_{il}^k \in {\bm \Psi}_i^k } n_l exp({{z}_{il}^k  }) }
   | {z}_{ij}^k \in {\bm \Psi}_i^k
   \}.
\end{equation}

\subsection{Nested Feature Learning}
%After obtaining ${\bf p}^* ({\bf x}_i;{\bm \theta}_k)$, 
The nested feature learning means that the feature learning is conducted on both the global view of all categories and the partial view of some important categories.
Following this idea, the nested feature learning is conducted on the individual learning as well as the intra-expert and inter-expert collaborative learning.
%We can conduct the individual learning from both the global view and partial view, which are constructed on all categories and those selected hard categories, respectively.
For the individual learning, formulating the learning in a nested way helps the network
achieve a global and robust learning on all categories, 
and also enhances the meticulous distinguishing ability to distinguish the input sample from 
the most confusing categories.
To be specific, the loss on the selected important categories for the balanced individual learning
is denoted as:
\begin{equation}\label{eq_bil_P}
   L_{bil}^{\mathcal{P}} = -\textstyle{\sum\nolimits_k} log( { {\bf p}}^{*}_{y_i} ({\bf x}_i;{\bm {\bm \theta} }_k) ).
\end{equation}
The whole loss of the balanced individual learning is the sum of the global part $L_{bil}^{\mathcal{G}}$ and partial part $L_{bil}^{\mathcal{P}}$, which is illustrated as the following formula:
\begin{equation}\label{eq_bil_all}
   L_{bil} = L_{bil}^{\mathcal{G}} + L_{bil}^{\mathcal{P}}.
\end{equation}

The intra-expert and inter-expert collaborative learning also can be conducted from the global view on all categories and  a partial view on the selected important categories. 
For InterCL, the corresponding distillation on the selected important categories can be written as:
\begin{equation}\label{inter_dis_imp}
   L_{inter}^{\mathcal P} = 
   \frac{2}{K(K-1)} {\sum_{k}^{K} \sum_{q \neq k}^{K} }
   KL( {\bf p}^{*} ({\bf x}_i;{\bm \theta}_k) || {\bf p}^{*} ({\bf x}_i;{\bm \theta}_q) ).
\end{equation}
Similarly, the whole loss of the InterCL is as following:
\begin{equation}\label{inter_dis_all}
   L_{inter} = L_{inter}^{\mathcal G} + L_{inter}^{\mathcal P}.
\end{equation}
The term $L_{inter}^{\mathcal G}$ transfers the global and structured knowledge from each all categories,
and the other term $L_{inter}^{\mathcal P}$ focuses on some important categories in which the experts vary greatly in predictions. For the IntraCL, the distillation on some selected important categories can be formulated in a similar way, and we denote it as $L_{intra}^{\mathcal P}$. Therefore, the IntraCL 
with a nested form can be formulated as:
\begin{equation}\label{intra_dis_all}
   L_{intra} = L_{intra}^{\mathcal G} + L_{intra}^{\mathcal P}.
\end{equation}

\subsection{Model Training}
The overall loss in our proposed method consists of three parts:
the loss $L_{bil}$ for learning each expert individually to enhance model's discriminative capability,
the loss $L_{intra}$ and $L_{inter}$ for cooperation among different augmented images and different experts, respectively. The overall loss $L$ is formulated as:
\begin{equation}\label{mcl_loss}
   L =   L_{bil} + \lambda_1 L_{intra}  +  \lambda_2 L_{inter} 
\end{equation}
where $\lambda_1$ and $\lambda_2$ denote the weighting coefficients for IntraCL and InterCL, respectively.
Considering both IntraCL and InterCL are collaborative learning, we set their coefficients to the same
and denote it as $\lambda$ (i.e., $\lambda = \lambda_1 = \lambda_2$).
%For $L_{bil}$ and $L_{con}$,  they play their part inside the single expert, and we equally set their weighs as 1 in consideration of generality.
Those losses are optimized cooperatively and concurrently, which enhances the discriminative capability of the network and reduces the uncertainties in predictions.
Moreover, our method is essentially a multi-expert framework, and therefore, 
the prediction in the test stage can be obtained by each single network or an ensemble of all of them,
one for high efficiency and one for high performance.

\section{Experiments}

\subsection{Datasets and Protocols}
We conduct experiments on five widely used datasets,
%for long-tailed visual recognition, 
including CIFAR10-LT~\cite{cui2019class}, 
CIFAR100-LT~\cite{cui2019class}, 
ImageNet-LT~\cite{liu2019large},
Places-LT~\cite{zhou2017places},
and iNaturalist 2018~\cite{van2018inaturalist}. 
%Following the common evaluation protocol\cite{zhang2021bag,kang2019decoupling,cui2021parametric},
%the model are trained on the long-tailed training set, and evaluated on the uniform validation or test set.
%Detailed descriptions about employed datasets are clarified in the following.
%As for Protocols,  following\cite{zhang2021bag,kang2019decoupling,cui2021parametric}, we train the model on the long-tailed distribution training set and evaluate their performance on the corresponding uniform test set.

\textbf{CIFAR10-LT and CIFAR100-LT}~\cite{cui2019class} are created from 
the original balanced CIFAR datasets~\cite{krizhevsky2009learning}.
%with the number of samples decaying exponentially across different classes.
Specifically, the degree of data imbalance in datasets is controlled by the use of an Imbalance Factor (IF),
which is defined by dividing the number of the most frequent category by that of the least frequent category.
The imbalance factors of 100 and 50 are employed in these two datasets.
\textbf{ImageNet-LT}~\cite{liu2019large} is sampled from the popular ImageNet dataset~\cite{deng2009imagenet} under long-tailed setting following the Pareto distribution with power value $\alpha$=6. 
ImageNet-LT contains 115.8K images from 1,000 categories.
%and the number of images in each category ranges from 5 to 1280.
\textbf{Places-LT} is created from the large-scale dataset Places~\cite{zhou2017places}.
This dataset contains 184.5K images from 365 categories.
%and the number of each classes ranges from 5 to 4,980.
\textbf{iNaturalist 2018}~\cite{van2018inaturalist} is the largest dataset for long-tailed visual recognition.
%The images in this dataset are collected in the wild scenarios for species identification of animals and plants.
iNaturalist 2018 contains 437.5K images from 8,142 categories, and it is extremely imbalanced with an imbalanced factor of 512.

According to previous works~\cite{cui2019class,kang2019decoupling}
the top-1 accuracy is employed for evaluation.
Moreover, for iNaturalist 2018 dataset, 
we follow the works~\cite{cai2021ace,kang2019decoupling} to divide classes into many (with more than 100 images), medium (with 20 $\sim$ 100 images) and few (with less than 20 images) splits,
and further report the results on each split. 

\subsection{Implementation Details}
%\textcolor{blue}{
For CIFAR10/100-LT, following~\cite{cao2019learning, zhang2021bag}, we adopt ResNet-32~\cite{he2016deep} as our backbone network and liner classifier for all the experiments. Input images are randomly cropped with size 32$\times$32.
%and horizontal flipped with  the probability of 0.5. 
We utilize ResNet-50~\cite{he2016deep}, ResNeXt-50~\cite{xie2017aggregated} as our backbone network for ImageNet-LT, ResNet-50 for iNaturalist 2018 and pretrained ResNet-152 for Places-LT respectively, based on~\cite{liu2019large,kang2019decoupling,cui2021parametric}. Following~\cite{zhang2021distribution}, the cosine classifier is utilized for these models. Following previous works~\cite{zhou2020bbn,kang2019decoupling,cui2021parametric}, we resize the input image to 256$\times$256 pixels and take a 224$\times$224 crop from the original image or its horizontal flip. We use the SGD with a momentum of 0.9 and a weight decay of $2 \times 10^{-4}$ as the optimizer to train all the models.
As for experiments on CIFAR10/100-LT, the initial learning rate is 0.1 and decreases by 0.1 at epoch 320 and 360, respectively. The learning rate for Places-LT is 0.02 and decreases by 0.1 at epoch 10 and 20.
For the rest datasets, the initial learning rate is set to 0.2 and decays by a cosine scheduler to $1 \times 10^{-4}$.
Unlike the previous works PaCo~\cite{cui2021parametric} and NCL~\cite{li2022nested}, we utilize less training epochs for ImageNet-LT and iNaturalist 2018, which is 200.
And the training epochs for Places-LT is 30, same as previous works~\cite{cui2021parametric,kang2019decoupling}.
%The models of CIFAR-LT are trained on a single NVIDIA RTX3090 GPU with the batch size of 128. 
Cumulative gradient~\cite{tan2022cross} is utilized for ImageNet-LT and iNaturalist 2018. Specifically, the training batch size is set to 128, but the gradient keeps accumulating and the parameters are updated every two epochs.
This trick uses less GPU memory to achieve similar training results as batch size of 256.
Two experts are utilized for all datasets.
Two augmented copies are utilized for ImageNet-LT and four for the rest.
In addition, for fair comparison, following~\cite{cui2021parametric}, RandAugment~\cite{cubuk2020randaugment} is also used for all the experiments. The influence of RandAugment will be discussed in detail in the Sec.~\ref{Component_Analysis}. 
These models are trained on 8 NVIDIA Tesla V100 GPUs.
%For the sake of generality, the loss ratio $\lambda_1$ and $\lambda_2$, which play their part inside the single expert, are set to 1. 
%The temperature $\tau$ in self-supervised loss is set to 0.2 and the size of queue N is set to 65536.
The $\beta = C_{hard} / C$ in HCM is set to 0.3, and the coefficient of InterCL and IntraCL loss $\lambda$ is set to 0.6. The influence of $\beta$ and $\lambda$ will be discussed in detail in the Sec.~\ref{Component_Analysis}.
%}

\setlength{\tabcolsep}{5pt}
\begin{table}[t]
\centering
\caption{Comparisons on CIFAR100-LT and CIFAR10-LT datasets with the IF of 100 and 50. $^\dagger$ indicates the ensemble performance is reported.
%Bold indicates the best.
}
\begin{threeparttable}
\resizebox{0.5\linewidth}{!}{
\begin{tabular}{l|c|cc|cc}
\toprule[1pt]
\multirow{2}{*}{Method} & \multirow{2}{*}{Ref.} & \multicolumn{2}{c|}{ CIFAR100-LT } & \multicolumn{2}{c}{ CIFAR10-LT } \\ 
\cline{3-6} 
%\midrule[0.5pt]
 &  & 100  & 50 & 100 & 50 \\ 
 \hline
%Focal loss~\cite{lin2017focal} &  & 37.4 & 42.4 & 70.4 & 75.3\\
%Mixup~\cite{zhang2017mixup} &  & 39.5 & 45.0 & 73.1 & 77.8\\
CB Focal loss~\cite{cui2019class} & CVPR'19 & 38.7 & 46.2 & 74.6 & 79.3\\ 
LDAM+DRW~\cite{cao2019learning} & NeurIPS'19 & 42.0 & 45.1 & 77.0 & 79.3\\
LDAM+DAP~\cite{jamal2020rethinking} & CVPR'20 & 44.1 & 49.2 & 80.0 & 82.2\\
BBN~\cite{zhou2020bbn} & CVPR'20 & 39.4 & 47.0 & 79.8 & 82.2\\
LFME~\cite{xiang2020learning} & ECCV'20 & 42.3 & -- & -- & --\\
CAM~\cite{zhang2021bag} & AAAI'21 & 47.8 & 51.7 & 80.0 & 83.6\\
Logit Adj.~\cite{menon2020long} & ICLR'21 & 43.9 & -- & 77.7 & --\\
Xu et al.~\cite{xu2021towards} & NeurIPS'21 & 45.5 & 51.1 & 82.8 & 84.3 \\ 
LDAM+M2m~\cite{kim2020m2m}& CVPR'21 & 43.5 & -- & 79.1 & --\\
MiSLAS~\cite{zhong2021improving} & CVPR'21 & 47.0 & 52.3 & 82.1 & 85.7 \\
LADE~\cite{hong2021disentangling} & CVPR'21 &  45.4 & 50.5 & -- & --\\
Hybrid-SC~\cite{wang2021contrastive} & CVPR'21 & 46.7 & 51.9 & 81.4 & 85.4 \\
DiVE~\cite{he2021distilling} & ICCV'21 & 45.4 & 51.3 & -- & -- \\
SSD~\cite{li2021self_iccv} & ICCV'21 & 46.0 & 50.5 & -- & --\\
PaCo~\cite{cui2021parametric} & ICCV'21 & 52.0 & 56.0 & -- & --\\
%OTLM+RIDE~\cite{peng2021optimal} & ICLR'22 & 51.4 & -- &-- & --\\
xERM~\cite{zhu2021cross} & AAAI'22 & 46.9 & 52.8 & -- & -- \\
RISDA~\cite{chen2021imagine} & AAAI'22 & 50.2 & 53.8 & 79.9 & 84.2\\
Batchformer~\cite{hou2022batchformer} & CVPR'22 & 52.4 & -- & -- & --\\
%WD~\cite{alshammari2022long} & CVPR'22 & 53.4 & 57.7 & -- & --\\
RIDE (4 experts)$^\dagger$~\cite{wang2020long} & ICLR'21 & 49.1 & -- & -- & --\\
ACE (4 experts)$^\dagger$~\cite{cai2021ace}& ICCV'21 & 49.6 & 51.9 & 81.4 & 84.9\\
TLC (4 experts)$^\dagger$~\cite{li2021trustworthy} & CVPR'22 & 49.8 & -- & 80.4 & --\\
\hline
NCL~\cite{li2022nested} & CVPR'22 & 53.3 & 56.8 & 84.7 & 86.8 \\
NCL (3 experts)$^\dagger$~\cite{li2022nested} & CVPR'22 & 54.2 & 58.2 & 85.5 & 87.3 \\
\hline
BSCE (baseline) & -- & 50.6 &   55.0    &   84.0    &   85.8 \\
NCL++  & -- &  \textbf{54.8} &   \textbf{58.2}    &   \textbf{86.1}    &   \textbf{88.0} \\
NCL++ (2 experts)$^\dagger$  & -- & \textbf{56.3} &  \textbf{59.8}    &   \textbf{87.2}    &   \textbf{88.8} \\
\bottomrule[1pt]
\end{tabular}}
%\begin{tablenotes}  
%        %\footnotesize       
%        \item The ensemble on $^{\dagger}$three experts and $^{\ddagger}$two experts.
%\end{tablenotes}
\label{results_cifar}
\end{threeparttable} 
\end{table}

\setlength{\tabcolsep}{5pt}
\begin{table}[t]
\centering
\caption{Comparisons on ImageNet-LT and Places-LT datasets. $^\dagger$ indicates the ensemble performance is reported.
%Bold indicates the best.
}
\resizebox{0.5\linewidth}{!}{
\begin{tabular}{l|c|ccc|c}
\toprule[1pt]
\multirow{2}{*}{Method} & \multirow{2}{*}{Ref.}  & \multicolumn{3}{c|}{ImageNet-LT} &  Places-LT  \\ 
\cline{3-6} 
%\midrule[0.5pt]
 &  & Res10 & Res50 & ResX50 & Res152\\ 
\hline
%Focal loss~\cite{lin2017focal} & \\
OLTR~\cite{liu2019large} & CVPR'19              & 34.1 & -- & -- & 35.9\\
BBN~\cite{zhou2020bbn} & CVPR'20                &   & 48.3 & 49.3 & --\\
%Logit Adj.~\cite{menon2020long} & \\
%\hline
%OLTR～\cite{liu2019large} & \\
NCM~\cite{kang2019decoupling} & ICLR'20         & 35.5 & 44.3 & 47.3 & 36.4\\
cRT~\cite{kang2019decoupling} & ICLR'20         & 41.8 & 47.3 & 49.6 & 36.7\\
$\tau$-norm~\cite{kang2019decoupling} & ICLR'20 & 40.6 & 46.7 & 49.4 & 37.9\\
LWS~\cite{kang2019decoupling} & ICLR'20         & 41.4 & 47.7 & 49.9 & 37.6\\
BSCE~\cite{ren2020balanced} & NeurIPS'20        & --  & -- & -- & 38.7 \\
%LDAM+DRW~\cite{cao2019learning} & \\
%LFME~\cite{xiang2020learning} & \\
%LDAM+M2m~\cite{kim2020m2m}& \\
%CAM~\cite{zhang2021bag} & \\
Xu et al.~\cite{xu2021towards} & NeurIPS'21 & 42.9 & 48.4 & -- & -- \\
DisAlign~\cite{zhang2021distribution} & CVPR'21 & --  & 52.9 & -- & --\\
DiVE~\cite{he2021distilling} & ICCV'21          & --  & 53.1 & -- & --\\
SSD~\cite{li2021self_iccv} & ICCV'21            & --  & --& 56.0 & --\\
PaCo~\cite{cui2021parametric} & ICCV'21         & --  & 57.0 & 58.2 & 41.2\\
ALA Loss~\cite{zhao2021adaptive} & AAAI'22      & --  & 52.4 & 53.3 & 40.1 \\
xERM~\cite{zhu2021cross} & AAAI'22              & --  & -- & 54.1 & 39.3  \\
RISDA~\cite{chen2021imagine} & AAAI'22          & --  & 50.7 & -- & --\\
MBJ~\cite{liu2020memory} & AAAI'22              & --  & -- & 52.1 & 38.1 \\
WD~\cite{alshammari2022long} & CVPR'22          & --  & 53.9 & -- & -- \\
BatchFormer~\cite{hou2022batchformer} & CVPR'22 & 47.6 & 57.4 & -- & 41.6 \\
RIDE (4 experts)$^\dagger$~\cite{wang2020long} & ICLR'21              & --  & 55.4 & 56.8 & --\\
MBJ+RIDE (4 experts)~\cite{liu2020memory} & AAAI'22         & --  & -- & 57.7 & -- \\
ACE (3 experts)$^\dagger$~\cite{cai2021ace} & ICCV'21                 & 44.0 & 54.7 & 56.6 & --\\
TLC (4 experts)$^\dagger$~\cite{li2021trustworthy} & CVPR'22          & --  & 55.1 & -- & --\\
\hline
NCL~\cite{li2022nested}  & CVPR'22      & 46.8  &57.4& 58.4 &  41.5\\
NCL (3 experts)$^\dagger$~\cite{li2022nested} & CVPR'22    & 47.7  &59.5& 60.5 & 41.8 \\
\hline
BSCE (baseline) & --  & 45.7 &53.9 &53.6&40.2\\
NCL++ & -- & \textbf{49.0} & \textbf{58.0} & \textbf{59.1} & 42.0 \\
NCL++ (2 experts)$^\dagger$ & -- & \textbf{49.9} & \textbf{59.6} & \textbf{60.9} & \textbf{42.4} \\
\bottomrule[1pt]
\end{tabular}
}
\label{results_imagenet}
\end{table}

\setlength{\tabcolsep}{5pt}
\begin{table}[t]
\centering
\caption{Comparisons on iNaturalist 2018 dataset with ResNet-50. $^\dagger$ indicates the ensemble performance is reported.
%Bold indicates the best.
}
\resizebox{0.5\linewidth}{!}{
\begin{tabular}{l|c|ccc|c}
\toprule[1pt]
\multirow{2}{*}{Method} & \multirow{2}{*}{Ref.}  & \multicolumn{4}{c}{ iNaturalist 2018} \\ 
\cline{3-6} 
%\midrule[0.5pt]
 &   &  Many  & Medium & Few & All \\ 
 \hline
%Focal loss~\cite{lin2017focal} & \\
%\hline
OLTR~\cite{liu2019large} & CVPR'19 & 59.0 & 64.1 & 64.9 & 63.9\\
BBN~\cite{zhou2020bbn} & CVPR'20 & 49.4 & 70.8 & 65.3 & 66.3\\
DAP~\cite{jamal2020rethinking}& CVPR'20 & -- & -- & -- & 67.6\\
NCM~\cite{kang2019decoupling} & ICLR'20 & \\
cRT~\cite{kang2019decoupling} & ICLR'20 & 69.0 & 66.0 & 63.2 & 65.2\\
$\tau$-norm~\cite{kang2019decoupling} & ICLR'20 & 65.6 & 65.3 & 65.9 & 65.6\\
LWS~\cite{kang2019decoupling} & ICLR'20 & 65.0 & 66.3 & 65.5 & 65.9\\
LDAM+DRW~\cite{cao2019learning} & NeurIPS'19 & -- & -- & -- & 68.0\\
Logit Adj.~\cite{menon2020long} & ICLR'21 & -- & -- & -- & 66.4\\
%LFME~\cite{xiang2020learning} & \\
%LDAM+M2m~\cite{kim2020m2m}& \\
CAM~\cite{zhang2021bag} & AAAI'21 & -- & -- & -- & 70.9 \\
%Xu et al.~\cite{xu2021towards} & NeurIPS'21 &  \\
SSD~\cite{li2021self_iccv} & ICCV'21 & \\
PaCo~\cite{cui2021parametric} & ICCV'21 & -- & -- & -- & 73.2\\
RIDE (3 experts)$^\dagger$~\cite{wang2020long} & ICLR'21 & 70.9 & 72.4 & 73.1 & 72.6\\
ACE (3 experts)$^\dagger$~\cite{cai2021ace} & ICCV'21 & -- & -- & -- & 72.9\\
ALA Loss~\cite{zhao2021adaptive} & AAAI'22 & 71.3 & 70.8 & 70.4 & 70.7 \\
xERM~\cite{zhu2021cross} & AAAI'22 & -- & -- & -- & 67.3 \\
RISDA~\cite{chen2021imagine} & AAAI'22 & -- & -- & -- & 69.1\\
MBJ~\cite{liu2020memory} & AAAI'22 & -- & -- & -- &  70.0\\
MBJ+RIDE~\cite{liu2020memory} & AAAI'22 & -- & -- & -- & 73.2\\
WD~\cite{alshammari2022long} & CVPR'22 & 71.2 & 70.4 & 69.7 & 70.2 \\
BatchFormer~\cite{hou2022batchformer} & CVPR'22 & 65.5 & 74.5 & 75.8 & 74.1\\
\hline
NCL~\cite{li2022nested} & CVPR'22 &\textbf{72.0}&\textbf{74.9}& 73.8 &\textbf{74.2} \\
NCL (3 experts)$^\dagger$~\cite{li2022nested} & CVPR'22 &\textbf{72.7}&\textbf{75.6}& 74.5& 74.9 \\
\hline
BSCE (baseline) & -- &67.5&72.0&71.5&71.4 \\
NCL++ & -- & 71.6 & 73.9 & \textbf{74.7} & 74.0 \\
NCL++ (2 experts)$^\dagger$ & -- & 72.2 & 75.3 & \textbf{75.7} & \textbf{75.2}  \\
\bottomrule[1pt]
\end{tabular}
}
\label{results_inatu}
\end{table}

\subsection{Comparisons to Prior Arts}

We compare the proposed NCL++ with previous state-of-art methods, like BBN~\cite{zhou2020bbn}, RIDE~\cite{wang2020long}, NCL~\cite{li2022nested}.
Besides, the baseline result of single network with using BSCE loss is also reported.
Two experts is the default setting for NCL++, which has smaller model size than NCL, but achieves better performance.
Comparisons on CIFAR10/100-LT are shown in Table~\ref{results_cifar},
comparisons on ImageNet-LT and Places-LT are shown in Table~\ref{results_imagenet},
and comparisons on iNaturalist 2018 are shown in Table~\ref{results_inatu}.
Our proposed method achieves the state-of-the-art performance on all datasets.
For only using a single expert for evaluation, 
our NCL++ outperforms previous methods on CIFAR10-LT, CIFAR100-LT,
ImageNet-LT and Places-LT with accuracies of 
86.1\% (IF of 50), 54.8\% (IF of 100), 58.0\% (with ResNet-50) and 42.0\%, respectively.
Compared with NCL~\cite{li2022nested}, the proposed NCL++ achieves significant improvements on CIFAR-LT (54.8\% vs. 53.3\%) and ImageNet-LT with ResNet10 (49.0\% vs. 46.8\%).
When further using an ensemble for evaluation, 
the performance on CIFAR10-LT, CIFAR100-LT,
ImageNet-LT, Places-LT and iNaturalist2018
can be further improved to 87.2\% (IF of 50),
56.3\% (IF of 100), 59.6\% (with ResNet-50), 42.4\%
and 75.2\%, respectively.
\textbf{
NCL++ uses fewer experts than NCL (2 experts vs. 3 experts), but achieves better ensemble performance on all the datasets.
}
Similar to NCL, a single network can be used for evaluation, which will not bring extra computation but still outperforms previous multi-experts method, like ACE, TLC, etc..
When using ensemble of two experts, the performance has been further significantly improved.

\subsection{Component Analysis}
\label{Component_Analysis}
\begin{figure}[t]
\centering
    {\includegraphics[width=0.5\linewidth]{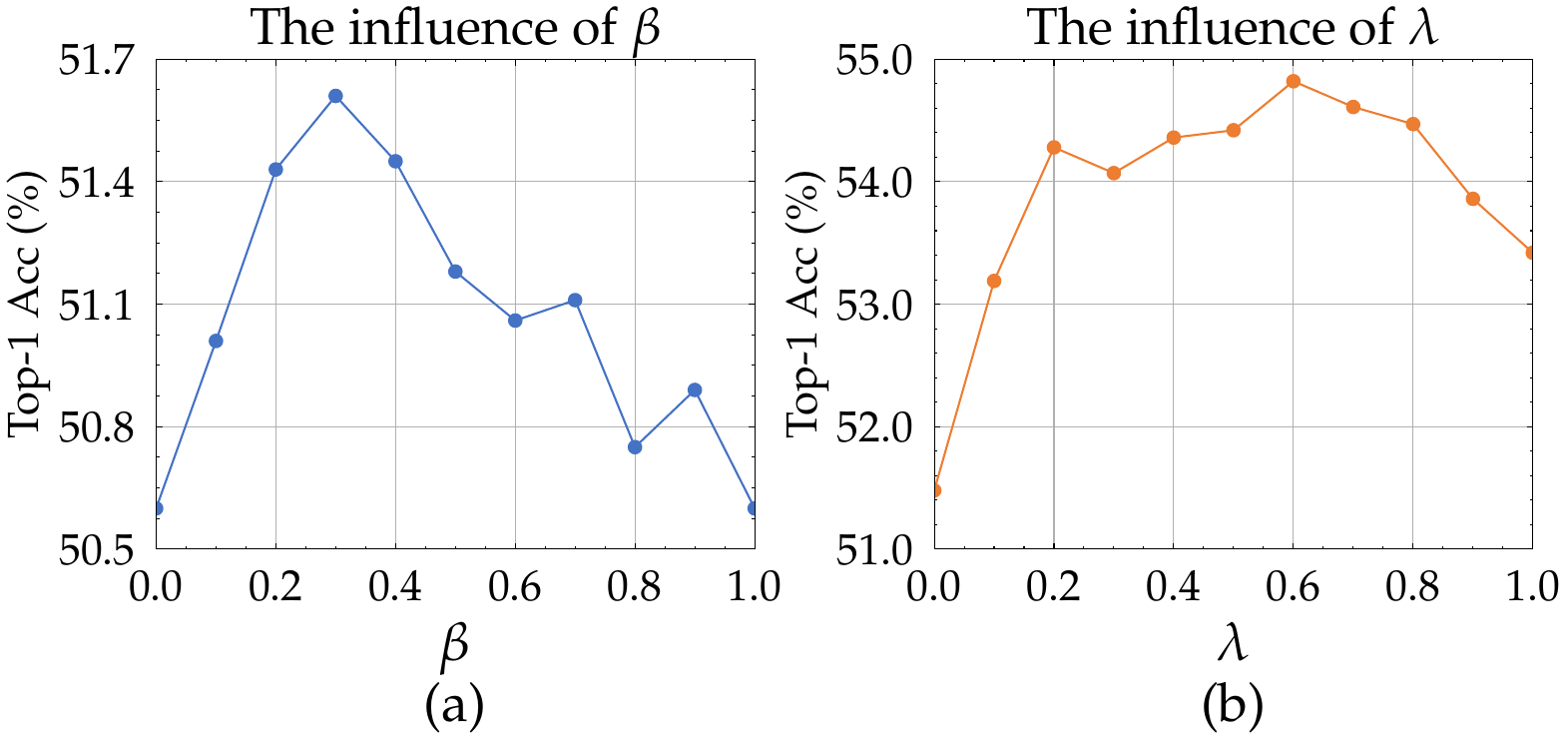}}
    %\vspace{-0.6cm}
    \caption{ Parameter analysis of (a) the ratio $\beta$ and (b) the loss weight $\lambda$ on 
    CIFAR100-LT dataset with IF of 100.
    %(a) shows the experimental results of using various $\beta$
    %(a) comparisons of diferent $\beta$ on CIFAR100-LT dataset with imbalance ratio of 100. Let $\beta$ donate the value of $\frac{C_{hcf}}{C_{all}}$, the network achieves the best performance when $\beta = 0.3$, i.e., selecting 30 hard categories for CIFAR100-LT.
    }
    \label{fig_hcf}
\end{figure}

\textbf{Influence of the ratio of hard categories.}
%To enhance network's discriminating ability,
%HCF pays attention to a small number of categories that are hard to be distinguished.
The ratio of selected hard categories is defined as $\beta = C_{hard} / C$.
Experiments on our BIL model 
are conducted within the range of $\beta$ from 0 to 1 as shown in Fig.~\ref{fig_hcf} (a).
%and the corresponding experimental results are show in Fig.~\ref{fig_hcf}.
The highest performance is achieved when setting $\beta$ to 0.3.
%The performance first increases and then decreases along with the increase of $\beta$,
%and the highest performance is achieved when setting $\beta$ to 0.3.
Setting $\beta$ with a small and large values brings limited gains
due to the under and over explorations on hard categories.
%Specifically, $\beta=0.0$ means the baseline network with a BSCE loss, where HCF is not adopted. 
%The network of $\beta=1.0$ achieves similar performance with that of $\beta=0.0$,
%because $\beta=1.0$ also means classifying on all categories, 
%which is equal to the network of using BCES loss twice.

\textbf{Effect of loss weight.}
To search an appropriate value for $\lambda$,
experiments on the proposed NCL++ with a series of $\lambda$ 
are conducted as shown in Fig.~\ref{fig_hcf} (b).
$\lambda$ controls the contribution of 
knowledge distillation among multiple experts and augmented images in total loss.
The best performance is achieved when $\lambda=0.6$,
which shows a balance is achieved between individual learning and collaborative learning.
Intuitively, the network achieves marginal improvements with a small or a large $\lambda$,
which shows that both ignoring and over emphasizing the collaborative learning are not optimal.

\begin{figure}[!t]
\centering
    {\includegraphics[width=0.5\linewidth]{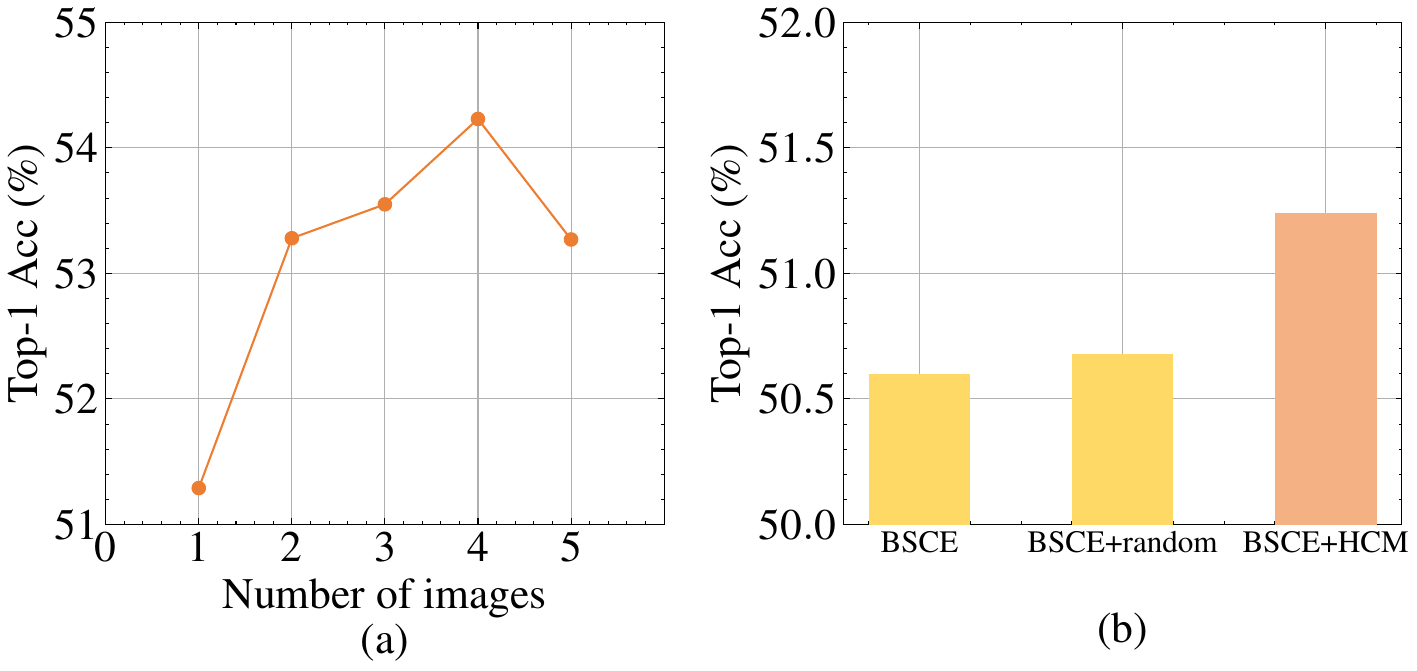}}
    \vspace{-0.3cm}
    \caption{ 
    (a) Comparisons of using different number of augmented image copies in IntraCL with a single network. (b) Comparisons of using hard categories selected by HCM or random categories.
    All experiments are conducted on CIFAR100-LT with an IF of 100.
    %'single' indicate the average valid accuracy of multi-experts, while the ensemble means the valid accuracy of average outpot logits of multi-experts.
    }
    \label{fig_num_hcm}
\end{figure}

\textbf{Impact of different number of augmented copies in IntraCL.}
As shown in Fig.~\ref{fig_num_hcm} (a), we take experiments for IntraCL with different number $T$ of augmented copies. The experiments are taken without using InterCL and thus only a single expert is employed. As we can see, the performance achieves the highest when taking four augmented copies for an image. 
Besides, $T=2$ is also a nice choice where the performance can be dramatically improved without a lot of extra computation.
%the performance is dramatically improved when setting $T=2$ compared with the .
%Therefore, $T=4$ is employed in the following experts.

\textbf{Hard categories vs. random categories.}
To further verify the effectiveness of the proposed HCM, 
we also take the experiments with random categories for comparisons as shown in Fig.~\ref{fig_num_hcm} (b) (denoted by 'BSCE + random'). 'BSCE + random' performs slightly better than the baseline method 'BSCE',
but its performance is still far worse than ‘BSCE + HCM’.
It shows that training on hard categories selected by the proposed HCM really helps to 
improve the discriminative capability of the model.

\textbf{Impact of different number of experts in InterCL.}
As shown in Fig.~\ref{fig_multi_expert}, experiments using different number of experts
are conducted. The ensemble performance is improved steadily as the number of experts increases,
while for only using a single expert for evaluation, 
its performance can be greatly improved when only 
using a small number of expert networks, e.g., two experts.
Therefore, two experts are mostly employed in our multi-expert framework for a balance 
between complexity and performance.

\textbf{Single expert vs. multi-expert.}
Our method is essentially a multi-expert framework,
and the comparison among using a single expert or an ensemble of multi-expert is a matter of great concern. As shown in Fig.~\ref{fig_multi_expert},
%and the performance of using different number of experts is a matter of great concern.
%We conduct experiments with increasing the number of experts from 1 to 7,
%and experimental results are shown in Fig~\ref{fig_multi_expert}.
%Thus, we conduct experiments of various expert numbers
%and report the performance on both a singe expert and an ensemble of all experts for comparisons.
%the single network performance and ensemble performance for comparisons.
%As shown in Fig.~\ref{fig_multi_expert}, 
%the single network performance is reported when only using a expert.
the ensemble can perform better than a single network on the performance over all classes and many splits.
%Nevertheless, for medium and few splits, the conclusion is interesting, 
%where the single network performance is higher than that of multi-expert ensemble.
%This may be due to the fact that there is great uncertainty of predictions on medium and few splits,
%and collaborative learning is a better choice than simply averaging the logits for eliminating the uncertainty.

\iffalse
\begin{figure}[!t]
\centering
    {\includegraphics[width=0.5\linewidth]{expert_num_ana.pdf}}
    \vspace{-0.3cm}
    \caption{ Comparisons of using different expert numbers on CIFAR100-LT with an IF of 100.
    We report the performance on both  a single network and an ensemble.
    Specifically, the performance on a single network is reported as the average accuracy on all experts,
    and the ensemble performance is computed based on the averaging logits over all experts.
    %'single' indicate the average valid accuracy of multi-experts, while the ensemble means the valid accuracy of average outpot logits of multi-experts.
    }
    \label{fig_multi_expert}
\end{figure}
\fi

\begin{figure}[!t]
\centering
    {\includegraphics[width=0.5\linewidth]{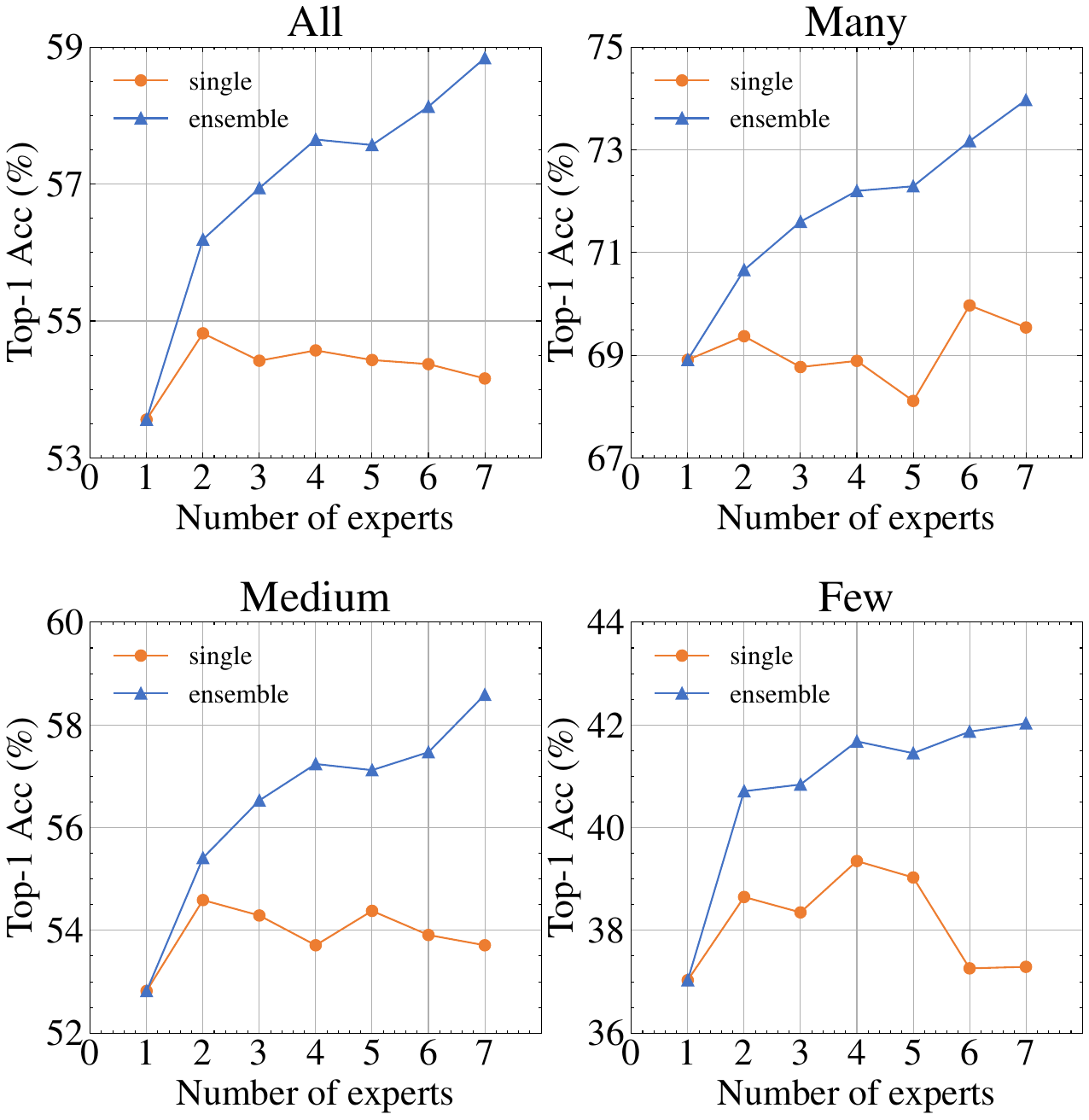}}
    \vspace{-0.3cm}
    \caption{ 
    %Comparisons of using different expert numbers in 'Inter' and different image numbers in 'Intra'.
    %'single' indicate the average valid accuracy of multi-experts, while the ensemble means the valid accuracy of average outpot logits of multi-experts.
    Comparisons of using different expert numbers of InterCL on CIFAR100-LT with an IF of 100.
    We report the performance on both  a single network and an ensemble.
    Specifically, the performance on a single network is reported as the average accuracy on all experts,
    and the ensemble performance is computed based on the averaging logits over all experts.
    }
    \label{fig_multi_expert}
\end{figure}

\begin{figure*}[!t]
\centering
    {\includegraphics[width=1.0\linewidth]{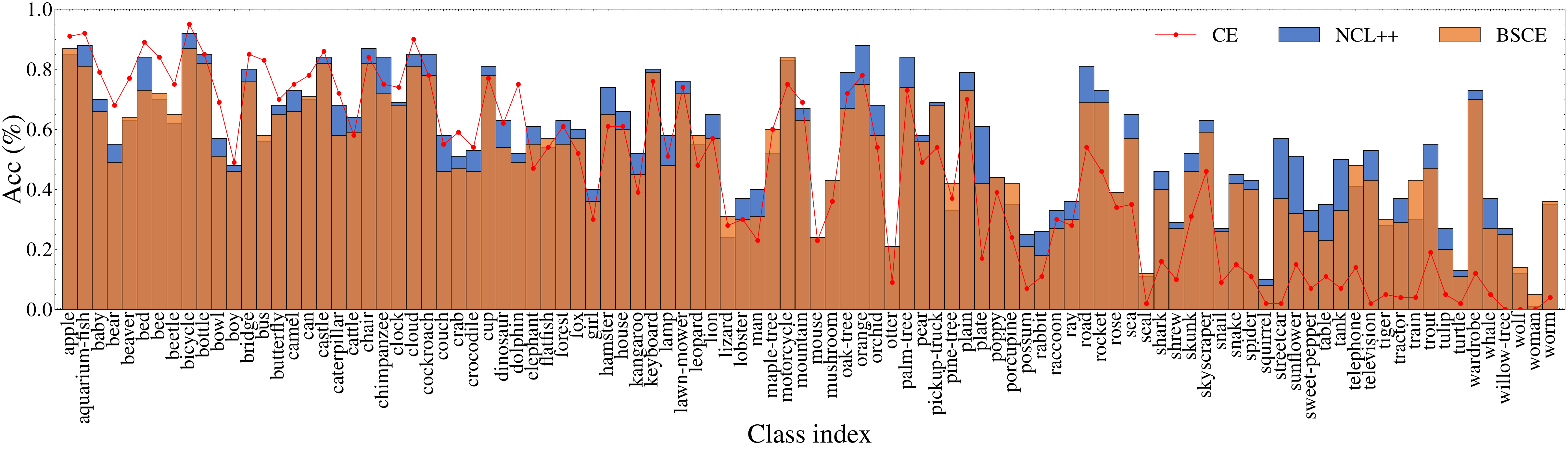}}
    \vspace{-0.3cm}
    \caption{ 
    %per-class accuracy on CIFAR100-LT-100.
    The accuracy of all categories on CIFAR100-LT with an IF of 100. 
    The horizontal axis represents the details of categories, 
    and the vertical axis represents their accuracies. 
    Note that we sort the categories according to the number of images, 
    where the number of images decreases from the left to right. 
    In other words, the leftmost and the rightmost indicate the category with the maximum and minimum number of
    images, respectively.
    }
    \label{fig_acc_each_catgorey}
\end{figure*}

\textbf{Influence of data augmentations.}
Data augmentation is a common tool to improve performance. 
For example, previous works~\cite{zhang2021bag,cai2021ace,zhong2021improving}
use Mixup~\cite{zhang2018mixup} %in addition PaCo~\cite{cui2021parametric} utilize 
and RandAugment~\cite{cubuk2020randaugment} to obtain richer feature representations. 
Our method follows PaCo~\cite{cui2021parametric}
to employ RandAugment~\cite{cubuk2020randaugment} for experiments.
%To investigate how much improvement this strategy can bring,
%experiments are conducted as shown in Fig.~\ref{auto_aug_analysis}.
As shown in Table~\ref{auto_aug_analysis},
the performance is improved by about 3\% to 5\% when employing RandAugment for training.
However, our high performance depends not entirely on RandAugment.
When using Mixup for augmentation, our method achieves the performance of 51.37\% and 53.38\% for single expert and ensemble of them respectively.
Even with the simple data augmentation (random crop, random horizontal flip), our method still performs very well, the ensemble model reaches an amazing performance of 51.57\%.
The results show the superiority of our method over state-of-the-art methods for both simple and complex data augmentation.

\setlength{\tabcolsep}{8pt}
\begin{table}[t]
\centering
\caption{Comparisons of training the network  with simple augmentation (SimAug), RandAugment (RandAug) and Mixup.
Experiments are conducted on CIFAR100-LT dataset with an IF of 100. $^\dagger$ indicates the ensemble performance is reported.
%Bold indicates the best.
}
\resizebox{0.5\linewidth}{!}{
\begin{tabular}{c|c|c|c}
\toprule[1pt]
%\multicolumn{5}{c|}{ Method } & Acc \\ 
%\cline{1-5} 
%\midrule[0.5pt]
 Method    & SimAug & RandAug & Mixup\\ 
% \midrule[0.5pt]
\hline
Current SOTA        &
49.1$^\dagger$~\cite{wang2020long}&   52.4~\cite{hou2022batchformer}  &  49.4$^\dagger$~\cite{cai2021ace} \\
\hline
CE                  &   41.68 & 44.79 & 41.85\\ 
BSCE                &   45.69 & 50.60 & 49.84\\
%BSCE+NCL           &   47.93 & 53.31 & 50.9\\
%BSCE+NCL$^\dagger$ &   49.22 & 54.42 & 52.3\\
BSCE+NCL++           & 48.76 & 54.82 & 51.37\\
BSCE+NCL++$^\dagger$ & 51.57 & 56.33 & 53.38\\
\bottomrule[1pt]
\end{tabular}
}
\label{auto_aug_analysis}
\end{table}

\textbf{Analysis on computation cost.}
We compare the computation cost between the proposed method and other state-of-the-art (SOTA) methods as shown in Table~\ref{cost_analsysis}.
For a fair comparison, we set two experts equally for both NCL and NCL++.
Due to using multiple networks for collaborative learning, NCL and NCL++ have more computation costs in the training stage. 
%Note that NCL++ has less computation cost than NCL, where the self-supervision module is replaced with the newly proposed IntraCL module. 
In the test stage, the NCL and its improved version, NCL++, only uses a single network for evaluation and it contains less computation cost but with a higher accuracy compared with other SOTA methods, like NCM~\cite{kang2019decoupling} and TCL~\cite{li2021trustworthy}. 
Moreover, compared with NCL++, NCL contains more computations (with higher GFLOPs) while a lower performance accuracy (57.4\% vs. 58.0\%). As we know, the only difference between NCL and NCL++ is that the self-supervision module is replaced with the newly proposed IntraCL module. Obviously, by replacing the self-supervision part with IntraCL, the computation cost is largely reduced while the performance is improved.

\setlength{\tabcolsep}{12pt}
\begin{table}[t]
\centering
\caption{
Comparisons of computation cost between the proposed method and other state-of-the-art methods. The analysis is conducted on ImageNet-LT dataset with the network of ResNet-50.
}
\resizebox{0.5\linewidth}{!}{
\begin{threeparttable} 
\begin{tabular}{c|c|c|c}
\toprule[1pt]
%\multicolumn{5}{c|}{ Method } & Acc \\ 
%\cline{1-5} 
%\midrule[0.5pt]
  \multirow{2}{*}{Method}   & \multicolumn{2}{|c|}{ FLOPs (G) } & \multirow{2}{*}{Accuracy (\%)} \\ 
  \cline{2-3} 
     & Train & Test & \\ 
% \midrule[0.5pt]
\hline
NCM~\cite{kang2019decoupling}  & 4.11  &  4.11 & 44.3 \\
%RIDE (2 experts)  & 4.52   & 3.71 & 54.4 \\
TLC~\cite{li2021trustworthy} (3 experts) & 6.55 & 6.55 & 55.1\\
\hline
NCL~\cite{li2022nested} (3 experts) & 17.03 & 4.11$^*$ & 57.4\\ 
NCL++ (2 experts) & 8.24 & 4.11$^*$ & 58.0\\ 
\bottomrule[1pt]
\end{tabular}
\begin{tablenotes} 
    \item $^*$ evaluating the model with a single expert.
\end{tablenotes} 
\end{threeparttable} 
}
\label{cost_analsysis}
\end{table}

\textbf{Fixed vs. learnable loss weights.}
In this paper, we fixedly set the loss weights $\lambda_1$ and $\lambda_2$ 
and select the appropriate values for them by a cross-validation (see Fig.~\ref{fig_hcf}). The loss weights also can be learn automatically as opposed to the fixed values. In this sub-section, we set loss weights $\lambda_1$ and $\lambda_2$ as learnable parameters and compare it with the fixed manner.
Following the work~\cite{zhang2020self}, we normalize the two learnable values $\lambda_1$ and $\lambda_2$ by a softmax function before using them to aggregate the losses, which ensures that $\lambda_1$ and $\lambda_2$  are non-negative values and $\lambda_1 + \lambda_2 = 1$. The experimental results are shown in Table~\ref{learnable_weight}. As we can see, compared to the fixed weights, using the learnable loss weights can obtain a higher accuracy on the train set but receive a lower accuracy on the test set.
It shows that the learnable loss weights may lead to overfitting and setting the parameters as fixed values may achieve better generalizations.

\setlength{\tabcolsep}{12pt}
\begin{table}[t]
\centering
\caption{
Comparisons of using fixed and learnable loss weights. The experiments are conducted on CIFAR100-LT dataset with an IF of 100.
}
\resizebox{0.5\linewidth}{!}{
\begin{tabular}{c|c|c}
\toprule[1pt]
%\multicolumn{5}{c|}{ Method } & Acc \\ 
%\cline{1-5} 
%\midrule[0.5pt]
Method   & Acc. on train set & Acc. on test set  \\ 
% \midrule[0.5pt]
\hline
Fixed   & 76.7 & 54.8\\
Learnable  & 82.3 & 54.5\\
\bottomrule[1pt]
\end{tabular}
}
\label{learnable_weight}
\end{table}

\textbf{Ablations studies on all components.}
In this sub-section, we perform detailed ablation studies for our NCL++ on CIFAR100-LT dataset,
which is shown in Table~\ref{analysis_ablation}. 
To conduct a comprehensive analysis, we evaluate the proposed components 
including Balanced Individual Learning (BIL), IntraCL, InterCL and the Nested Feature Learning on hard categories (denoted as 'NFL$_{hard}$'). 
%The BIL is indeed a plain network with taking BSCE as the loss function, and it can be regarded the baseline of the proposed NCL++. 
The ablation studies start from a naive method, which does not take any one of above components.
To be specific, the naive method is indeed a plain network with taking the Cross-Entropy (CE) loss as the loss function.
%Besides, we also present the performance of a naive method without taking any one of above components. This 
Compared with the naive method, the BIL improve the performance from 44.79\% to 50.60\%, 
which is a considerable improvement. 
This is because the naive method does not consider anything about the imbalanced data distribution.
For the BIL, it dramatically increases the contributions of tail classes while suppressing that of head classes,
which could greatly improve the performance by picking low hanging fruit.
For other components like IntraCL, InterCL and NFL$_{hard}$, 
all of them can steadily improve the performance. 
For example, based on BIL, the proposed InterCL and IntraCL improve the performance by 2.53\% and 3.34\%, respectively.
When both InterCL and IntraCL are employed, this improvement can be enlarged to 3.76\%.
Then, the NFL$_{hard}$ also gains the improvements from 54.36\% to 54.82\%.
Finally, we take the model ensemble of all experts (two experts are employed) 
to produce the final predictions, and the performance is further improved to 56.33\%.
The steadily performance improvements clearly show the effectiveness of the proposed NCL++.

\setlength{\tabcolsep}{3pt}
\begin{table}[t]
\centering
\caption{
Ablation studies on CIFAR100-LT dataset with an IF of 100.
'BIL', 'InterCL' and 'IntraCL' represent the individual learning or collaborative learning of the global view on all categories. For 'NFL$_{hard}$', it means adding the nested feature learning of the partial view on hard categories.
'Ensemble' means that the final predictions are obtained by averaging the results on multiple experts.
In the proposed NCL++, two experts are employed. 
%'SS' indicates self-supervision. 'BOD$_{all}$' and 'BOD$_{hard}$' represent the balanced online distillation on all categories and only hard categories, respectively.
%NBOD means the setting when both 'BOD$_{all}$' and 'BOD$_{hard}$' are employed.
%Experiments are conducted on the framework of containing three experts.
%$^\dagger$ indicates three experts are employed in our framework.
%Bold indicates the best.
}
\resizebox{0.5\linewidth}{!}{
\begin{tabular}{ccccc|cc}
\toprule[1pt]
%\multicolumn{5}{c|}{ Method } & Acc \\ 
%\cline{1-5} 
%\midrule[0.5pt]
BIL          & InterCL    &  IntraCL   &  NFL$_{hard}$ &  Ensemble  & Acc. \\
% \midrule[0.5pt]
\hline
             &            &            &             &             & 44.79 \\
\checkmark   &            &            &             &             & 50.60 \\
\checkmark   &            &            & \checkmark  &             & 51.24 \\
\checkmark   &\checkmark  &            &             &             & 53.13 \\
%\checkmark   &\checkmark  &            & \checkmark  &             & \textcolor{red}{52.69}\\
\checkmark   &            & \checkmark &             &             & 53.94 \\
%\checkmark   &            & \checkmark & \checkmark  &             & \textcolor{red}{53.56}\\
\checkmark   &\checkmark  & \checkmark &             &             & 54.36\\
\checkmark   &\checkmark  &\checkmark  & \checkmark  &             & 54.82\\
\checkmark   &\checkmark  &\checkmark  & \checkmark  & \checkmark  &56.33      \\
\bottomrule[1pt]
\end{tabular}
}
%\vspace{-0.3cm}
\label{analysis_ablation}
\end{table}

\begin{figure}[t]
\centering
    {\includegraphics[width=0.6\linewidth]{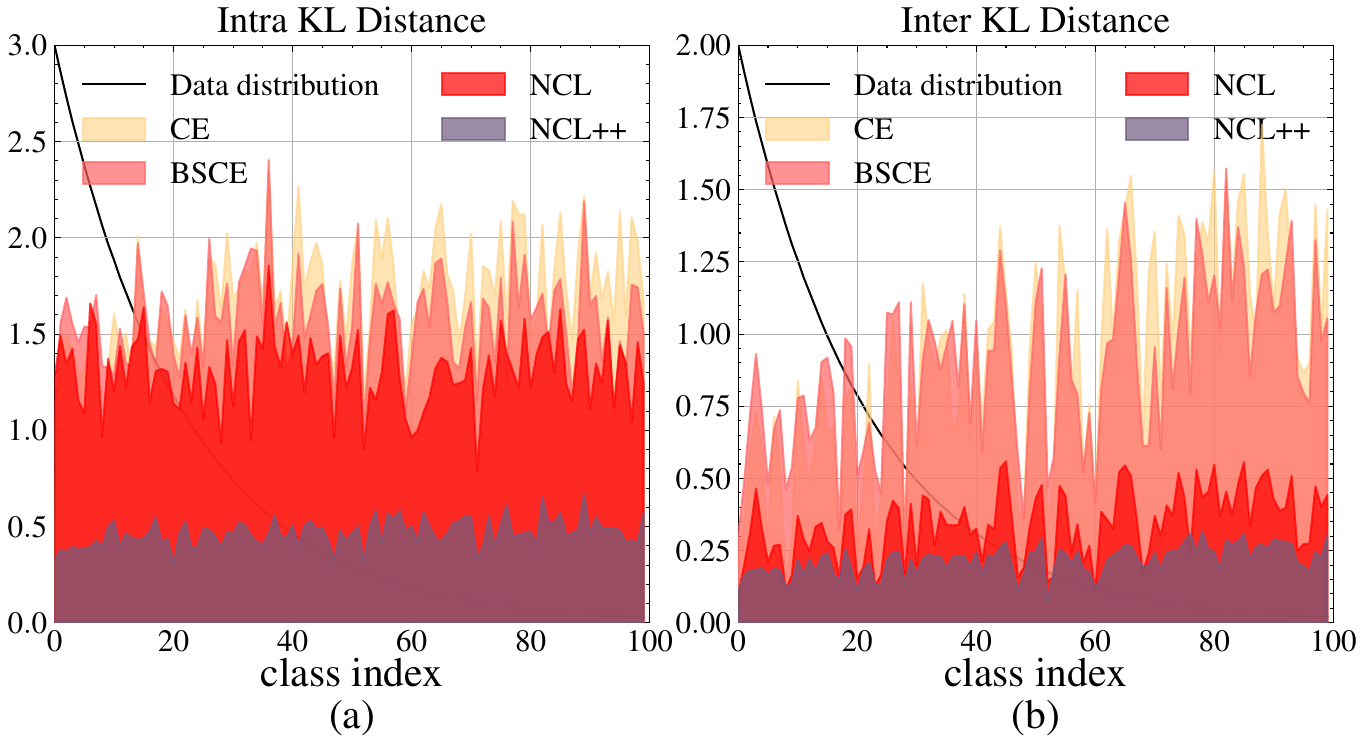}}
    % \vspace{-0.8cm}
    \caption{ 
    %(a) The distribution of the largest softmax probability of hardest negitive category. 
    %(b) The average KL distance between two models' output probabilities on the test set. 
    The average KL distance between (a) the output probabilities of two experts with respect to the same image
    and (b) the output probabilities of two augmented image copies with respect to the same expert.
    Analysis is conducted on CIFAR100-LT with an IF of 100. Best viewed in color.
    }
    \label{fig_top10}
\end{figure}

\subsection{Discussion and Further Analysis}

\textbf{KL distance of pre/post collaborative learning.}
To analyze the uncertainty of pre/post collaborative learning,
we visualize the average KL distance of two different augmented copies (see Fig.~\ref{fig_top10} (a)) or two experts (see Fig.~\ref{fig_top10} (b)) with CE, BSCE, NCL and NCL++. 
As we can see, both CE and BSCE do not consider the uncertainty in long-tailed learning,
and thus the corresponding learned experts still show large KL distances among different experts and different image augmented copies. For the NCL, it only considers the learning uncertainty among different experts.
Therefore, the KL distance between different image augmented copies is still large in NCL, which shows the intra-expert uncertainty has not been reduced. 
For the proposed NCL++, the KL distances between different experts and different image augmented copies are greatly reduced, which shows that both the intra- and inter-expert uncertainties are effectively alleviated.
%Besides, the KL distance is more balanced than that of BSCE and CE, which indicates that collaborative learning is of help to the long-tailed bias reduction.

\begin{figure}[t]
\begin{minipage}{0.4\linewidth}
\centerline{\includegraphics[width=1\linewidth]{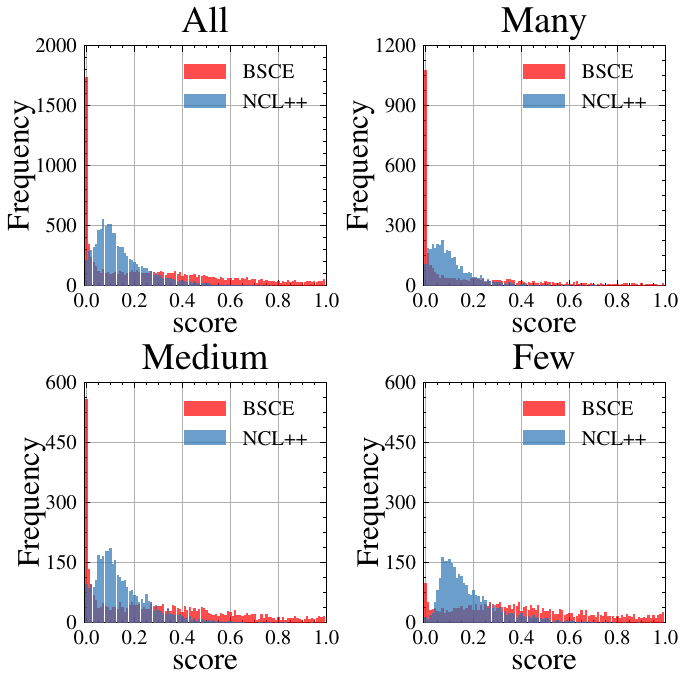}}
\centerline{(a)}
\end{minipage}
%\ 
%\hspace{-10mm}
%\begin{minipage}{0.33\linewidth}
%\centerline{\includegraphics[width=1\linewidth]{imagenet_res50_acc_image.pdf}}
%\centerline{(b) ImageNet-LT}
%\end{minipage}
%\
\ \ \ \ \ \ \ \ 
\begin{minipage}{0.4\linewidth}
\centerline{\includegraphics[width=1\linewidth]{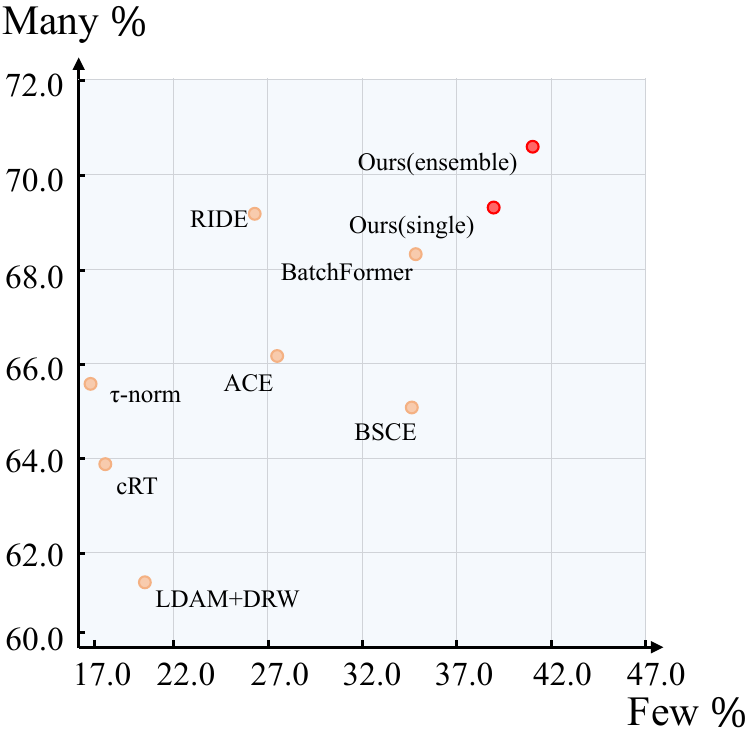}}
\centerline{(b) }
\end{minipage}
\caption{
(a) The distribution of the probability of hardest negative category.
(b) Comparisons of our proposed method and some representative methods over many and few splits.
Experiments are conducted on CIFAR100-LT with an IF of 100. Best viewed in color.
}
\label{fig_compare}
\end{figure}

%%%%%%%%%%%%%%%%%%%%%%%%
\textbf{Score distribution of hardest negative category.}
Deep models normally confuse the target sample with the hardest negative category.
Here we visualize the score distribution for the baseline method ('BSCE') 
and our method ('NCL++') as shown in Fig~\ref{fig_compare} (a).
The higher the score of the hardest negative category is, the more likely it is to produce false recognition.
The scores in our proposed method are mainly concentrated in the range of 0-0.4,
while the scores in the baseline model are distributed in the whole interval (including the interval with large values). This shows that our NCL++ can considerably reduce the confusion with the hardest negative category.

\begin{figure}[t]
\centering
    {\includegraphics[width=0.6\linewidth]{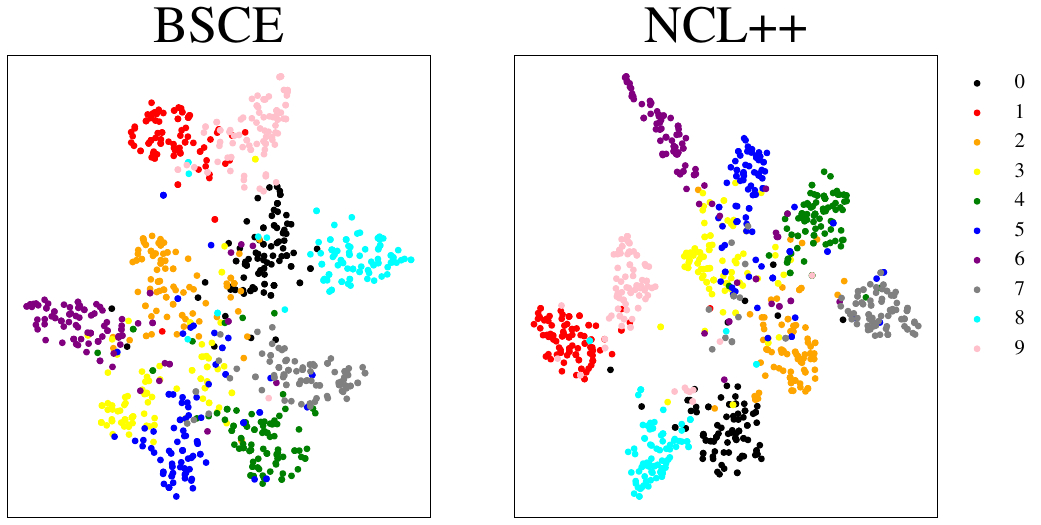}}
    %\vspace{-0.8cm}
    \caption{ 
    t-SNE visualization between BSCE and NCL++ on CIFAR10-LT with an IF of 100.
    Different colors indicate different categories.
    }
    \label{fig_tsne}
\end{figure}

\textbf{t-SNE visualizations.}
We use the t-SNE~\cite{van2008visualizing} to visualize the features of the baseline method BSCE and the proposed NCL++. 
%As shown in the Fig.~\ref{fig_tsne}, we can observe that our NCL++ generates more compact features than BSCE (e.g., the red color). It means the features extracted by the proposed NCL++ stay closer and thus being more reliable for visual recognition.
%Based on our DHAA, the extracted features with the same age stay closer and thus being more reliable for facial age estimation.
Compared to the baseline BSCE, the two-dimensional t-SNE map of the proposed NCL++ seems to be better clustered as shown in Fig.~\ref{fig_tsne}, (e.g., the red color). It shows that the proposed NCL++ generate more compact feature representation than the baseline BSCE, where the features of the same category extracted by the network stay closer. The nicely clustered and compact features indicate the model are well-trained for long-tailed visual recognition, which can achieve better classification and recognition performance.

\textbf{Experiments on balanced datasets.}
Although NCL++ is proposed based on the phenomenon of long-tailed datasets,
except for the re-balanced part, the rest components of NCL++ can still be applied to classification tasks on balanced datasets,
such as CIFAR~\cite{krizhevsky2009learning}.
We conduct experiments on CIFAR100 as shown in Table~\ref{results_original_cifar}.
Cross Entropy (denoted as 'CE') is utilized as the baseline method and the result of the single expert is reported.
As we can see from the results, the proposed NCL++ achieves considerable improvement over the baseline.
The performance is improved by 2.61\% with ResNet-10 and 3.20\% with ResNet-32.
It is undeniable that even in balanced datasets there are still confusing categories and the uncertainty of model training will still exist.
In these scenarios, NCL++ also shows its advantages.

\setlength{\tabcolsep}{16pt}
\begin{table}[h]
\centering
\caption{Comparisons on CIFAR100 dataset.
%Bold indicates the best.
}
\begin{threeparttable}
\resizebox{0.5\linewidth}{!}{
\begin{tabular}{l|c|c}
\toprule[1pt]
Method & ResNet-10  &   ResNet-32\\ 
\hline
CE      & 64.45  & 73.68\\
NCL     & 64.79  & 74.24\\
NCL++   & 67.06  & 76.88\\
\bottomrule[1pt]
\end{tabular}}
\label{results_original_cifar}
\end{threeparttable} 
\end{table}

% \setlength{\tabcolsep}{16pt}
% \begin{table}[h]
% \centering
% \caption{The comparison of computational complexity between different methods.
% %Bold indicates the best.
% }
% \begin{threeparttable}
% \resizebox{0.75\linewidth}{!}{
% \begin{tabular}{l|c|c|c}
% \toprule[1pt]
% Method & Train-FLOPs(G)  & Train-FLOPs(G) & Acc.\\ 
% \hline
% NCM~\cite{kang2019decoupling}	      & 4.110	& 4.110	& 44.3\\
% RIDE~(2 expert)~\cite{wang2020long}	  & 4.520	& 3.710	& 54.4\\
% TLC~(3 expert)~\cite{li2021trustworthy}&6.550	& 6.550	& 55.1\\
% NCL~(2 expert)~\cite{li2022nested}	   &17.03	& 4.110	& 57.4\\
% NCL++~(2 expert)	& 8.240	& 4.110	& 58.0 \\

% \bottomrule[1pt]
% \end{tabular}}
% \label{results_original_cifar}
% \end{threeparttable} 
% \end{table}

%%%----add by tan

\textbf{Analysis on all categories.}
As shown in Fig.~\ref{fig_acc_each_catgorey}, we present the accuracy on all categories for three methods,
namely the proposed method NCL++, the naive method CE and the baseline method BSCE.
Compared with the naive method CE, the proposed method NCL++ can dramatically improve the performance of tail classes,
which mainly benefits from the balanced learning (including balanced individual learning and balanced online distillation).
For example, for the sixth category from the last, the accuracy is improved from about 13\% to 74\%, where over 60\% is improved. Moreover, compared with the baseline method BSCE, 
our NCL++ can improve the performance on almost all categories,
which clearly shows the effectiveness of the proposed collaborative learning and NFL. 
%the accuracy of the naive method CE and the proposed method NCL++ on all categories are presented. The proposed method method dramatically improve the performance of tail classes. 
%For example, for the sixth from the last, the accuracy is improved from about 13\% to 74\%, where over 60\% is improved.

\textbf{Visual comparisons to prior arts.}
As shown in Fig.~\ref{fig_compare}, we compare the proposed method with prior arts on many (with more than 100 images) and few (with less than 20 images) splits. 
The comparisons on CIFAR100-LT are shown in Fig.~\ref{fig_compare} (b). As we can see, 
the proposed method achieves remarkable improvements on both many and few splits compared with previous SOTAs (e.g., RIDE and ACE). We take the comparisons on CIFAR100-LT as an example.
Many previous methods could perform well on many split (69.2\% for RIDE) but perform poorly on few split (only 26.3\% for RIDE). For the proposed NCL++, it dramatically improves the accuracy of the few split to 38.7\% with a single network through collaborative learning.

\iffalse
\setlength{\tabcolsep}{6pt}
\begin{table}[t]
\centering
\resizebox{0.5\linewidth}{!}{
\begin{tabular}{c|ccccc|c}
\toprule[1pt]
%\multicolumn{5}{c|}{ Method } & Acc \\ 
%\cline{1-5} 
%\midrule[0.5pt]
 balanced probability    &  Single expert       &  Ensemble \\ 
% \midrule[0.5pt]
 \hline
            &    52.1       &  53.4     \\ 
\checkmark  &   53.3        &  54.4     \\
\bottomrule[1pt]
\end{tabular}
}
\caption{
Comparisons of with or without using the balanced probability in NBOD.
Experiments are conducted on  CIFAR100-LT dataset with an IF of 100. 
}
\label{analysis_re-balancing}
\end{table}
\fi

\begin{figure}[t]
\centering
    {\includegraphics[width=0.5\linewidth]{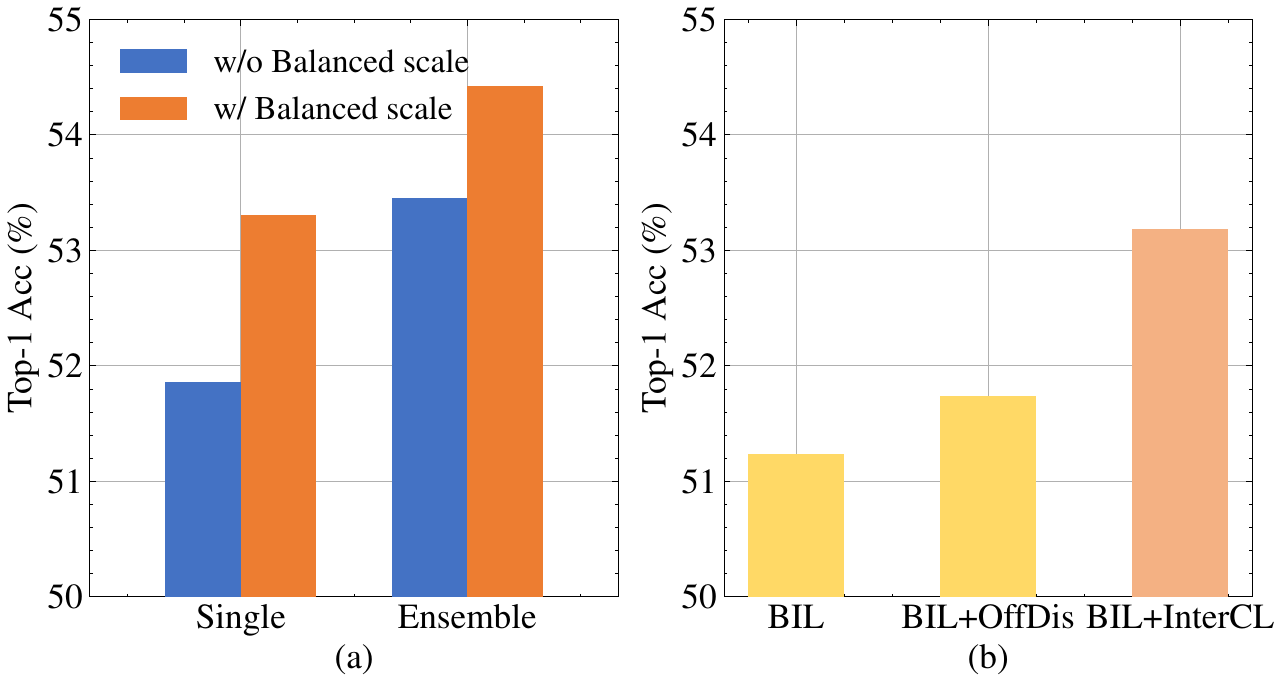}}
    \vspace{-0.3cm}
    \caption{ (a) Comparisons of whether using balanced scale in InterCL.
    (b) Comparisons of using offline distillation or our InterCL. Analysis is conducted on CIFAR100-LT with an IF of 100.
    }
    \label{balance_offdis}
\end{figure}

\iffalse

\textbf{NBOD without balancing probability.}
As shown in Fig.~\ref{balance_offdis} (a), 
when removing the balancing probability in NBOD (denoted as 'NOD')
both the performance of a single expert and a ensemble
decline about 1\%, which manifests the importance of employing the balanced probability
for the distillation in long-tailed learning.
%the performance of a single expert declines from 53.3\% to 52.1\%,
%and that of the ensemble also slides from 54.4\% to 53.4\%.

\textbf{Offline distillation vs. NBOD.}
To further verify the effectiveness of our NBOD, we employ a offline distillation for comparisons.
The offline distillation (denoted as 'NIL+OffDis') 
first employs three teacher networks of NIL to train individually, 
and then produces the teacher labels by using the averaging outputs over three teacher models.
The comparisons are shown in Fig.~\ref{balance_offdis} (b).
Although NIL+OffDis gains some improvements via a offline distillation, 
but its performance still 1.5\% worse than that of NIL+NBOD.
It shows that our NBOD of the collaborative learning 
can learn more knowledge than offline distillation.
\fi

\textbf{InterCL without balancing probability.}
As shown in Fig.~\ref{balance_offdis} (a), 
when removing the balancing probability in InterCL (denoted as 'w/o balanced scale')
both the performance of a single expert and an ensemble
decline about 1\%, which manifests the importance of employing the balanced probability
for the distillation in long-tailed learning.
%the performance of a single expert declines from 53.3\% to 52.1\%,
%and that of the ensemble also slides from 54.4\% to 53.4\%.

\textbf{Offline distillation vs. InterCL.}
To further verify the effectiveness of our InterCL, we employ an offline distillation for comparisons.
The offline distillation (denoted as 'BIL+OffDis') 
first employs three teacher networks of NIL to train individually, 
and then produces the teacher labels by using the averaging outputs over three teacher models.
The comparisons are shown in Fig.~\ref{balance_offdis} (b).
Although BIL+OffDis gains some improvements via an offline distillation, 
but its performance still 1.5\% worse than that of BIL+InterCL.
It shows that our InterCL of the collaborative learning
can learn more knowledge than offline distillation.

\textbf{Performance vs. data scale.}
We have conducted more experiments on ImageNet-LT to explore the relationship between performance and dataset size. As shown in Table.~\ref{performance_scale}, the performance will increase as the dataset scale increases. Compared to baseline method (BSCE), NCL++ consistently shows improvements across different dada scales. For example, our NCL++ improves the performance by 3.60\% when only using 20\% data, and improves the performance by 3.30\% when using all data. 
%This shows that there is no significant correlation between the effectiveness of our method and the data scale. 

\setlength{\tabcolsep}{12pt}
\begin{table}[t]
\centering
\caption{
The impact of training data with different ratios. Experiments are conducted on 
 ImageNet-LT with ResNet-10.
}
\resizebox{0.4\linewidth}{!}{
\begin{tabular}{c|c|c}
\toprule[1pt]
%\multicolumn{5}{c|}{ Method } & Acc \\ 
%\cline{1-5} 
%\midrule[0.5pt]
Data ratio	& Baseline Acc. (\%)& NCL++ Acc. (\%) \\ 
% \midrule[0.5pt]
\hline
20\%  &	40.64 &	44.24 (+3.60) \\
40\%  &	44.04 &	47.93 (+3.89) \\
60\%  & 44.66 &	48.05 (+3.39) \\
80\%  &	44.91 &	48.70 (+3.79) \\
100\% & 45.70 &	49.00 (+3.30) \\
\bottomrule[1pt]
\end{tabular}
}
\label{performance_scale}
\end{table}

\setlength{\tabcolsep}{12pt}
\begin{table}[t]
\centering
\caption{
Comparisons between our NCL++ and other contrastive learning methods.
Experiments are conducted on CIFAR100-LT with an IF of 100.
}
\resizebox{0.3\linewidth}{!}{
\begin{tabular}{c|c}
\toprule[1pt]
%\multicolumn{5}{c|}{ Method } & Acc \\ 
%\cline{1-5} 
%\midrule[0.5pt]
Method   & Acc. (\%) \\ 
% \midrule[0.5pt]
\hline
Hybrid-PSC~\cite{wang2021contrastive}  & 45.0 \\
PaCo~\cite{cui2021parametric}  & 52.0 \\
%\hline
NCL~\cite{li2022nested} & 53.3\\
NCL++ (ours) & 54.8 \\
\bottomrule[1pt]
\end{tabular}
}
\label{compare_contrastive}
\end{table}

\textbf{Comparisons to contrastive learning methods.}
There are some similarities between our method and contrastive learning methods. For contrastive learning, it aims to narrow the distance between positive pairs (augment copies from the same image) by contrastive learning losses. For our NCL++, it employs the KL loss to minimize the distance between positive sample pairs. Specifically, the definition of the positive sample pairs are two folds: one is the different augment copies from the same image in our IntraCL, and the other is the different predictions of the same image from different networks in InterCL. However, there are also some differences among them. For example, contrastive learning is an unsupervised method while our method is a supervised method. It is precisely because of supervised learning where each sample has a certain label, we further propose the nested feature learning to increase the network’s ability to distinguish the sample from the negative hard categories. We have compared the performance of our method and other contrastive learning based methods (including Hybrid-PSC~\cite{wang2021contrastive} and PaCo~\cite{cui2021parametric}) as shown in Table~\ref{compare_contrastive}. As we can see, our method outperforms them by a considerable margin. Regarding the aspect of hard sample mining, Hybrid-PSC enhances training on the tail by reweighting, while PaCo enhances training on the tail by resampling. These approaches mine hard samples based on categories. While in our method, we proposed HCM to focus on each individual sample, extracting hard categories specific to that sample. Therefore, our method is still different from the hard example mining, and has not been proposed before. We further take a simple experiment to compare the performance of our IntraCL and a classic contrastive learning MoCo. Actually, when replacing the IntraCL with MoCo’s contrastive learning module, our NCL++ degenerates to the previous version NCL~\cite{li2022nested}. As shown in Table~\ref{compare_contrastive}, our IntraCL could achieve better performance, which shows the advantages of the proposed method.

\section{Conclusions}
In this work, we have proposed a Nested Collaborative Learning (NCL++)
to enhance the discriminative ability of the model in long-tailed learning by collaborative learning.
Five components including BIL, IntraCL, InterCL, HCM and NFL were proposed in our NCL++.
The goal of BIL is to enhance the discriminative ability of a single network.
IntraCL and InterCL conduct the collaborative learning to reduce the learning uncertainty.
For HCM and NFL, they aim to select hard negative categories and then formulate the learning in a nested way to improve the meticulous distinguishing ability of the model.
Extensive experiments have verified the superiorities of our method
over the state-of-the-art.

\textbf{Limitations and broader impacts.}
One limitation is that more GPU memory and computing power are needed
when training our NCL++ with multiple experts and augmented copies, which will bring higher training costs.
But fortunately, one expert is also enough to achieve promising performance in inference, which still shows some advantages compared to the ensemble-based multi-expert approaches. 
Moreover, the proposed method improves the accuracy and fairness of the classifier, which promotes the visual model to be further put into practical use. To some extent, it helps to collect large datasets without
forcing class balancing preprocessing, which improves efficiency and effectiveness of work. The negative impacts can
yet occur in some misuse scenarios, e.g., identifying minorities for malicious purposes. Therefore, the appropriateness
of the purpose of using long-tailed classification technology
is supposed to be ensured with attention.



%% References


\bibliographystyle{elsarticle-num}
\bibliography{main.bib}






% \end{thebibliography}

\end{document}